\documentclass[10pt,journal,compsoc]{IEEEtran}
\usepackage{graphicx}
\graphicspath{{./pdf/}{./jpeg/}{./graphic}{./figures/}}
\usepackage{amsmath}
\usepackage{amssymb}
\usepackage{bm}
\usepackage{amsfonts}
\usepackage{array}
\usepackage{pgf}
\usepackage{pgfplots}
\usepackage{tikz}
\usepackage{url}
\usepackage{ragged2e}
\usepackage{algorithm}
\usepackage[nocompress]{cite}
\usepackage{algorithmic}

\usetikzlibrary{calc,patterns,angles,quotes}
\usepackage{graphicx}
\ifCLASSOPTIONcompsoc
\usepackage[caption=false, font=normalsize, labelfont=sf, textfont=sf]{subfig}
\else
\usepackage[caption=false, font=footnotesize]{subfig}
\fi
\usepackage[capitalise]{cleveref}
\DeclareMathOperator{\E}{\mathbb{E}}

\DeclareMathOperator*{\argmin}{argmin}
\DeclareMathOperator*{\argmax}{argmax}
\DeclareMathOperator{\length}{Length}
\renewcommand{\raggedright}{\leftskip=0pt \rightskip=0pt plus 0cm}
\begin{document}

\title{Inferring Latent dimension of Linear Dynamical System with Minimum Description Length}
\author{Yang~Li
	\IEEEcompsocitemizethanks{\IEEEcompsocthanksitem The author is with the
		Department of Computer Science, University of Science and Technology of China,
		Anhui, CN, 230027.\protect\\ E-mail: csly@mail.ustc.edu.cn}
}

\IEEEtitleabstractindextext{
\raggedright{
\begin{abstract}

Time-invariant linear dynamical system arises in many real-world applications,
and its usefulness is widely acknowledged. A practical limitation with this
model is that its latent dimension that has a large impact on the model
capability needs to be manually specified. It can be demonstrated that a
lower-order model class could be totally nested into a higher-order class, and
the corresponding likelihood is nondecreasing. Hence, criterion built on the
likelihood is not appropriate for model selection. This paper addresses the
issue and proposes a criterion for linear dynamical system based on the
principle of minimum description length. The latent structure, which is omitted
in previous work, is explicitly considered in this newly proposed criterion. Our
work extends the principle of minimum description length and demonstrates its
effectiveness in the tasks of model training. The experiments on both univariate
and multivariate sequences confirm the good performance of our newly proposed
method.
\end{abstract} }

}

\maketitle
\IEEEdisplaynontitleabstractindextext
\IEEEpeerreviewmaketitle

\IEEEraisesectionheading{\section{Introduction}}
\label{sec:intro}

Time-invariant linear dynamical system (LDS) has been extensively used in
engineering and controlling the behaviors of physical systems. Because of its
mathematically analyzable structure and predictive behavior, many engineering
applications and physical systems can be accurately described by this model
\cite{Martens2010}. The LDS is proposed to model the statistical properties of
observable data and further seeks for a probabilistic explanation for the underlying generating
mechanism. It assumes that the observable data are correlated to the value of
finite underlying latent variables, or latent state which evolves over the time
course according to a linear transition equation. The unexplainable factors are
captured by the noise terms which are assumed to be Gaussian.

The usefulness of LDS is not limited to engineering. It has been extensively
applied in various domains. In statistical pattern recognition, LDS allows for a
formal approach to approximate the data source from which the observations are
assumed to be \emph{i.i.d.} sampled. This generative viewpoint coincides with
many model-based methodologies that carry out learning on alternative
representation, i.e. the model itself, rather than on the data directly. 
 
The exact inference within the LDS can be done via Expectation Maximum based
(EM-based) approaches. EM is an iterative optimization algorithm and converges
to a solution of maximum likelihood (ML) estimate of parameters. It is a greedy
heuristic solver, and consequently, it is sensitive to the
initialization when it is implemented on a multi-modal problem. Moreover, it
also has the drawback of converging to the boundary of its parametric space at
which the distributions of some variables degenerate to Dirac delta
functions. This happens often when the model is too complex for the given data and
thus resulting in meaningless estimates.

Another issue comes from the specification of latent space dimension. This issue
raises the usual trade-off between model capability and simplicity. With a large
latent space, we run the risk of over-fitting, while a small latent space
renders model rigid and less able to approximate the true underlying generating
mechanism. The above two issues are closely correlated in the sense that
EM-based approaches are more prone to get struck in a local maximum especially
in a high dimensional parametric space. Therefore, these two issues should not
be treated separately.

We start with an intuitive idea: all generative methods are about to finding and
regenerating regularities in data. The best model that could regenerate the
regularities is also the one that extracts most information from the data
\cite{Grunwald2004}. That is, the model, in some sense, provides a descriptor
that compresses the data most. This idea is formally expressed by the principle
of Minimum description length (MDL) \cite{Rissanen1978,Grunwald2004}. A model
that leads to the best compression of given data is chosen and is perceived as
generalizing well on unseen data. The main strength of MDL is that it can be
used for selecting the general form of a model as well as its parameters. 

However, unlike many standard statistical models, the latent state space in LDS
needs special attention. This particular structure increase the model capability
while makes it challenging to describe this model, which is necessary for
applying the principle of MDL. In this paper, we propose an inference criterion
especially for models with latent variables based on the MDL. We concentrate on
the real-valued linear dynamical system and demonstrate how minimum description
length could be utilized for selecting an LDS of suitable order without
increasing much computational cost. From our derivation, we show the deep
connection between system stability and code length. Based on this proposed
criterion, an algorithm which implements it selects a model that best suits the
observations while having low model complexity. This criterion could be easily
extended to other discrete cases without much efforts. By penalizing high
complex models, the likelihood function is actually bounded and thus the
EM-based approaches are less prone to the boundary of parametric space.

The rest of paper is organized as follows: In \cref{sec:background}, we give
background information that are relevant to our work. Specifically, in \cref{sec:lds}, we review
linear dynamical system as well as the EM algorithm in detail. In this part, we
will define notations that will be used throughout this study. The principle of
minimum description length is outlined in \cref{sec:mdl}. A special
emphasis is put on the idea behind this principle. Other work in the family of
panelized maximum likelihood criteria and how they are applied in model
selection are reviewed in \cref{sec:pmle}. In \cref{sec:learning}, we describe
the proposed criterion as well as an algorithm that is built
from EM. Section \ref{sec:exp} reports our
experiments and \cref{sec:conclusion} concludes the whole paper.

\section{Background}
\label{sec:background}

\subsection{Linear Dynamical System}
\label{sec:lds}

\begin{figure}	
\centering
\begin{tikzpicture}
 [L1Node/.style={circle,draw=black!40, fill=black!20,minimum size= 10mm},
 L2Node/.style={circle,draw=black!50,minimum size=10mm}, line width = 1 pt]
\node[L1Node](w1) at (2,0) {$\bm{y}_{1}$};
\node[L2Node](w2) at (2,-2) {$ \bm{x}_1 $};
\node[L1Node](w3) at (4,0) {$\bm{y}_{2}$};
\node[L2Node](w4) at (4,-2) {$ \bm{x}_2 $};
\draw[->] (w2) -- (w1);
\draw[->] (w2) -- (w4);
\draw[->] (w4) -- (w3);
\node[](split) at(5,-2) {$ \cdots $};
\node[L1Node](w8) at (6,0) {$\bm{y}_{t}$};
\node[L2Node](w7) at (6,-2) {$ \bm{x}_t $};
\node[L1Node](w5) at (8,0) {$\bm{y}_{t+1}$};
\node[L2Node](w6) at (8,-2) {$ \bm{x}_{t+1} $};
\draw[->] (w7) -- (w8);
\draw[->] (w6) -- (w5);
\draw[->] (w4) -- (w3);
\draw[->] (w7) -- (w6);
\end{tikzpicture}
\caption{The schematic diagram of Linear Dynamical System (LDS). In this
	diagram, the variables in shadow are observable while others need to be inferred.
	The observations are assumed to be linearly correlated with latent states, whose
	evolution over the time course specifies the statistical properties of
	observations.} \label{fig:lds}
\end{figure}
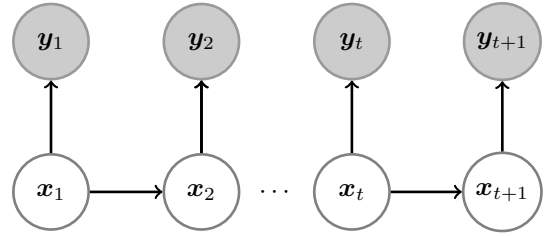

Many problems in engineering and controlling the physical systems involve the
real-valued multivariate sequential observations or time series. For such kind of data, the time-invariant
linear dynamical system (LDS) (aka. Kalman filter \cite{Kalman1960}) has been extensively used. To model the
statistical properties of data, the LDS correlates sequences to a
fixed-size latent variable vector or a finite-dimensional latent state, whose evolution over the sequential 
course makes up dynamics of data. In LDS, the state is assumed to be
in real domain and the noise terms are assumed to follow the Gaussian distribution.
Thus, the statistics properties could be explicitly expressed by a simple
formula, which is written as:
\begin{align}
\left\{
\begin{aligned}
\bm{x}_{t+1} &=& A\bm{x}_{t} + \bm{w}_t \\
\bm{y}_{t+1} &=& C\bm{x}_{t+1}+\bm{v}_{t+1}
\end{aligned}
\right.
\label{eq:lds}
\end{align}
where the set of real-valued vector $ \bm{x}_t \in \mathbb{R}^{d} $ and $
\bm{y}\in \mathbb{R}^{d_{out}} $ denote the latent variable and observation at 
time step $ t $, respectively; Transition matrix $ A $ is a coefficient matrix
that controls the evolution of latent states between two successive time steps;
$ C $ is often called observability matrix which specifies how observations are
generated from the present latent state. The initial density is also given as
Gaussian distribution with parameters $ \pi = \{\mu_0, R_0\} $. In LDS, the
noise terms $ u_t \in \mathbb{R}^{d} $ and $ v_{t+1} \in \mathbb{R}^{d_{out}} $
are assumed to follow zero-mean Gaussian distributions, whose covariances are
given as $ R_1 $ and $ R_2 $, respectively. Throughout this study, bold letters
are used to denote vectors and capitals for matrices of appropriate sizes. We drop
the sizes of matrices when the context is clear. We give a figure of LDS in
\cref{fig:lds}

For an LDS, one important aspect is its stability. An LDS is regarded to be
stable if all eigenvalues of the $ A $ are bounded up by one on magnitude. As can
be verified by empirical studies \cite{Boots2008,Huang2016}, if the transition
matrix $ A $ fails to satisfy this requirement, the model may generate unstable
output within which one component may diverges to infinity, thus causing significant distortion.
In the task of generating synthetic sequence or model simulation,
instability should be avoided if we need observe long correlated sequences. This condition is indispensable in both arithmetic formulation and many
real-world applications \cite{Huang2016,Saisan2001,Ravichandran2013}.

We reserve the symbols $ i $ and $ j $ to denote generic indexes, and $ t $
to indicate a time stamp. Let $ \bm{y}_t $ represent one particular sequence and
$ Y = [\bm{y}_1, \cdots, \bm{y}_t, \cdots, \bm{y}_k]^T \in
\mathbb{R}^{d_{out}\times k}$ be the collection of observations over time interval
$ t \in \{1,\cdots,k\} $. Likewise, let $ X $ be the collection of latent states
over the same interval, $ X = [\bm{x}_1, \cdots,\bm{x}_k]^T $. The likelihood 
of generating the set of data $ Y $ is concisely written as:

\begin{align}
\begin{split}
	p(Y|\theta) &= \int_X p(Y|X,\theta)p(d X,\theta)\\ 
&\propto \prod_t \int_{\bm{x}_t} p(\bm{y}_t|\bm{x}_t, \theta) p(d \bm{x}_t| \bm{x}_{t-1},\theta)
\end{split}
\end{align}
where $ \theta = \{A,C,d, R_1, R_2\} $ are the parameters waiting to be learned. In
particular, we include the latent dimension in this set, which is considered as a parameter rather than a prefixed quantity.

There are two principled ways in inferring the parameters from finite
observations. One involves only the likelihood function. The parameters are
inferred through solving a maximum likelihood (ML) problem expressed as below:
\begin{equation}
	\hat{\theta}_{ML} = \argmax\{\log p(Y|\theta)\}
\end{equation}

Another one makes use of probabilistic conjugate structure and places some prior
density function on the subset of $ \theta $. This approach selects a
parametric combination that maximize the \emph{posteriori} probability. It solves the following optimization problem as expressed:
\begin{equation}
	\hat{\theta}_{MAP} = \argmax \{\log p(Y|\theta) + \log p(\theta)\}
\end{equation}

However, closed-form solutions are not available for both the problems. When
the latent dimension $ d $ is known, the standard solver for these problems is the
EM algorithm \cite{Ghahramani1996,Dempster1977}, which executes in an iterative fashion finding a nondecreasing estimates 
for parameters. It is based on a simple intuition that interprets the
observations as incomplete data $ (Y,X) $. The missing part contains the information about
latent variables over the time course. This procedure is summarized in two steps
(the E-step and M-step). The EM solver first makes a guess about the
complete data $ (Y,X) $ and solves for the $ \theta $ that maximizes the
likelihood. Once we get an estimate for $ \theta $, it updates its belief on the
complete data and then iterates.

The convergence behavior of EM has been well studied \cite{Gupta2010}. Usually,
the EM estimate never gets worse than previous trails. It will find an estimate
that performs best locally. However, when the likelihood function has multiple
peaks or is multimodal, it is not guaranteed to find the global optimum. This
problem can be alleviated through random reinitialization, which starts EM from
multiple randomly chosen points and chooses the one with the best performance. A
more serious problem for an EM-based approach is that, when the likelihood
function is unbounded, it may converge to the boundary of the parameter space
where the covariance matrix of a noise term may becomes arbitrarily close to be
singular. When the model order is high, this phenomenon tends to happen
frequently. To alleviate this problem, a frequently used way is to impose soft
constraints on the covariance matrix \cite{Kloppenburg1997}. However, for model
selection where comparison is made among models of different orders, we need a
mixed strategy that can both handle the model selection while prevent the above
issues from happening.

\subsection{Minimum Description Length}
\label{sec:mdl}

The minimum description length criterion (MDL) \cite{Barron1991} is used for
compromising between model fitness and model complexity. The former is measured
via the goodness-of-fit which involves the likelihood of generating given data.
The trade-off between these two terms is necessary because a model cannot be
assessed solely on its fitness to the data, as more complex model can
necessarily lead to a high fitness yet may also overfit the data. A general idea
behind MDL is to choose a model which allows one to express the data with the
shortest possible \emph{description length}. As we will see, more complex model
necessarily needs long code to describe it and incurs long description length.
This principle coincides with the compression philosophy where any nonrandom
patterns can theoretically be used for a batter packing ratio.

When we are doing the model fitting, we are in fact trying to rebuild or approximate
the true data source by using the patterns and regularities existing in the
data. For non-random data, the model which truly generates the data actually provides the
most effective and parsimonious choice. From Shanon' coding theory \cite{Cover2012}, we are
able to construct a code based on this descriptor or the induced data
distribution from it, the resulting code length is the shortest theoretically.

When the true model is unknown, we need more description length for the
additional missing information. In addition to that, we also need to code the
model. These two parts make up the total description length. The chosen model
should summarize the data in the sense that the description length for the data,
as well as the model, should be close to the one that could be possibly achieved
with the true one. As no model could be more informative than the true model for
the data, a model in a particular model class $ \mathcal{M} $ is selected only
if it yields the shortest description length for the data and itself. This
criterion for model selection thus compromises between model fitness and model complexity.

Barron et. al. \cite{Barron1991} has shown that for sufficiently large sample
size, if the true model is on this countable list, then, with probability one, 
MDL could correctly identify the true model. The probability of making erroneous
selection goes to zeros as $ n\rightarrow \infty $. This result could also be
extended when the true model is not contained within the countable model list.
Provided that the true model could be approximated by a sequence of candidate
models in the sense of relative entropy. These inspiring researches provide support
for the usage of MDL in the model selection.

\subsection{Panelized Maximum Likelihood Estimation} 
\label{sec:pmle} 

The minimum description length could be viewed in a large family within which an approach appends a
nonnegative term appended to the likelihood estimation function to penalize a  high model
order or much model complexity. This term can be also viewed as a regularizer
that favors lower complex models. This class of methods is also known as
\emph{complexity penalized} or \emph{regularized maximum likelihood estimation}.

To start with, define $ M_d $ as the class of all possible models with latent
dimension $ d $. The ML criterion is not enough to estimate the model order
because $ M_k \subset M_{k+1} $ as one latent dimension could be set to zeros
and thus being inactive. That is, the lower dimensional class could be nested
totally to a higher order model class. A
straightforward way is to select from a range of $ k's $, which is likely to
contain the true one and check the model's fidelity to the data. The latent dimension
selected according to this deterministic principle is expressed as:
\begin{align}
\hat{d} = \argmax\{C[\theta(d),d], d\in \{d_{min}, \cdots, d_{max} \}\}
\end{align}
where $ C[\cdot,\cdot] $ is a particular selection criterion, and $ \theta(d) $
is the ML or MAP estimate provided that $ d $ is given beforehand. The commonly used
criterion is the probability of fitness or the \emph{a posteriori} probability
given the data. 

Different designs of the $ C[\cdot,\cdot] $ give rise to different model
selection criteria. The approaches following this methodology are deterministic
as they will lead to consistent model order estimates. The most popular choices
include: Akaike information criterion (AIC) \cite{Akaike1974}, Bayesian
information criterion (BIC) \cite{Raftery1995}, and the Fisher information
approximation (FIA) \cite{Gruenwald2007}
\begin{itemize}
	\item  Akaike information criterion (AIC)
	\begin{align}
	AIC = -2 \log p(Y|\hat{\theta}) + 2 n_\theta
	\end{align}
	\item  Bayesian information criterion (BIC)
	\begin{align}
	BIC = -2 \log p(Y|\hat{\theta}) +  n_\theta\log(n)
	\end{align}
	\item Fisher information approximation (FIA)
	\begin{align}
	\begin{split}
	FIA &= -\log p(Y|\hat{\theta}) + \frac{n_\theta}{2} \log \frac{n}{2 \pi} \\ 
	&+  \log \big ( \int_\theta \sqrt{\det \mathcal{I(\theta)}} d\theta \big )
	\end{split}
	\end{align}
\end{itemize}
where $ n $ denotes samples size; $ n_\theta $ is the number of free parameters;
$ \hat{\theta} $ is the ML or MAP estimate for the parametric setup; $
\mathcal{I(\theta)} $ is Fisher information matrix. $ p(Y|\theta) $ gives the
goodness-of-fit for a model under $ \hat{\theta} $. For all the three criteria,
the one which yields the lowest criterion value is considered the ``best'' model
in a list of candidate models.

Other approaches in model selection follow a different mechanism. They often
execute in a stochastic fashion. For example, Markov chain Monte Carlo (MCMC)
can either sample from a full posterior distribution with latent dimension $ d $
being unknown, or could implement with the assistance of various model selection
criteria. In terms of computational cost, these approaches are considered too
high to be useful in practical applications. These stochastic approaches, as
compared with deterministic criteria like AIC, BIC etc., would not lead to a
consistent estimator. Cross-validation and resampling are also useful in
estimating the model order. But their computational loads are closer to those of
stochastic approaches than those of some deterministic criteria listed above.

The previous works on MDL mostly concentrate on the static models, as an example,
Figueiredo et. al. explored MDL on a finite mixture model and proposed an
unsupervised learning scheme based on EM algorithm \cite{Figueiredo2002}. The investigation on
migrating the MDL to a dynamical model, especially the one with latent variables
is largely overlooked. In this study, we work on this problem and propose a
variant of EM algorithm. As we will demonstrate, the algorithm
turns out to be effective in approximating the true underlying model.
Furthermore, it also inherits the merits of avoiding the boundary of the
parameter space automatically, as already pointed out in \cite{Figueiredo2002}.

\section{Inference Criterion}
\label{sec:learning}

This section considers employing minimum description length (MDL) criterion for
selecting appropriate order for time-invariant linear dynamical system (LDS).
The overall procedure is a two-stage process that first estimates the effects
from truncating observability matrix and then considers the deviation brought by
using approximated latent states rather than the true ones. Different from the
common settings where MDL applies, LDS has latent variables or states that can
be inferred purely from observations and the approximated parameters. That is,
they are not totally independent. Therefore, they induce no increases on the
description length. We adopt this two-stage process, hoping to inspect two
different effects brought by parameters and latent variables in depth.

To begin with, suppose the parameters for an LDS are collectively denoted as $
\theta $ and they should be treated as a whole. We reserve $ i,j $ as index
variables whose ranges change over the context. The symbol $ T $ represents the
sample length or how long a sequence of observations is. The value $ N $
represents the sample size. We use different symbols for latent variables and
parameters as they will induce different treatments. Any specific value of the
state $ \bm{x} $ at time $ t $ is called state value, denoted by $ \bm{x}_t $.
When no index is attached, it should be viewed as a random variable vector.

Shannon' theory tells us that in a given model class, we can construct a code of
length $ \length[Y|\mathcal{M}] = \lceil \log p(Y|X,\mathcal{M}(\theta))\rceil$
for $ Y $. The logarithm is to the base of $ e $ \footnote{The natural logarithm
	measures information in the measure of \emph{nats}. If base 2 is used, the
	information is measured in bits. Both formulations can be conveniently
	transformed by a factor of $ \log2 $ and are equivalent in terms of
	minimization.}. Fortunately, we do not actually construct this code, only to know that it is
possible for the code to exist. For notational convenience, we will omit the $
\lceil \cdot \rceil $ in the rest of study.

Within the philosophy of MDL, it is not enough to encode the observations, we
need to encode the model as well. MDL states that the most appropriate choice
for model selection is the one that minimizes the total description length,
denoted as $ \operatorname{Length}(\mathcal{M},Y) $. To formulate this idea,
suppose the data are \emph{i.i.d.} generated from an unknown generator as $ Y =
\{\bm{y}_1,\bm{y}_2,\cdots,\bm{y}_n \} $, according to MDL, in a list of
candidate models $ \bigcup\limits_{d_{min}}^{d_{max}} \mathcal{M}(\theta(d)) $,
where $ d_{max} $ and $ d_{min} $ are upper and lower bound we shall consider.
The minimum complexity estimator $ \mathcal{M}(\hat{\theta})  $ is defined as
the one which achieves the minimum of the following formula:
\begin{align}
\length(Y,\mathcal{M}(\hat{\theta}))= \min_{\mathcal{M}(\theta)} \big \{ \underbrace{L(\mathcal{M}) }_{\length[\mathcal{M}]}+ \underbrace{\log \frac{1}{p(Y|\mathcal{M})}}_{\length[Y|\mathcal{M}]} \big \}
\end{align}

The above equation corresponds to the minimization of the total length of a
two-stage description of data. The first term could be interpreted as the
description length for a certain model, and the second term describes the
description length for the data. The usual trade-off happens here between models
and data. An oversimplified model approximates the true one in a coarse manner.
Thus, $ \length(\mathcal{M}(\hat{\theta})) $ could be small, but $
\length(p(Y|\mathcal{M}(\hat{\theta}))) $ could be large. The converse
also holds. A flexible model could approximate the true one to a high
accuracy, and the resulting $ \length(\mathcal{M}(\hat{\theta})) $
is large, but $ \length{p(Y|\hat{\theta})} $ can be small.

The general description of a model does not involve any structures. Any member
from a candidate set does not need to relate one another in any sense. To build the
description length for models, we require that $ \theta $ has a nice property in
an open set of its domain. Specifically, we require that $ \theta $ can define
a member in a model class unambiguously. That is, when $ \theta_1 \neq \theta_2
$, we have $ \mathcal{M}(\theta_1) \neq \mathcal{M}(\theta) $ in the sense of
any valid probabilistic disparity measures.

Within the methodology of MDL, the model parameters (specified by parameter $
\theta $) and the observations are quantized and encoded. The data themselves
could be real-valued has little effects on the inference process, as all actual
observations are truncated to some finite precision. The only necessary thing is
to replace the function $ p(Y|\theta) $ by $ p(Y|\theta) \eta^{d_{out}}  $,
where $ \eta^{n_{out}} $ is an operator truncating the output values. The
resulting code length becomes $ -\log p(Y|\theta) -d_{out} \log \eta $. The
quantization of observations to be finite precision affects the code length
through a term that is irrelevant to learning of parameters. Recall that in
practical applications, the precision is usually given in advance and
consequently $ -d_{out} \log \eta $ is an irrelevant term that does not affect
the overall process. In the following analysis, we can safely discard this term
without causing any difficulties.

As opposed to quantization of observations, which cause minor interest, the
truncation of parameters brings serious problem because the effects of
quantization can be amplified through the model and thus incur a loss on the
model capability. Consider a prior density $ p(\theta) $ for models and the
likelihood $ p(Y|\theta) $ for a given data $ Y $. Suppose the quantization on $
\theta $ is given by an operator $ \delta $. In the remaining, we should find a
way to evaluate the effects of the truncation. The main tool is the Taylor
series expansion.

Let $ \hat{\theta} $ be the quantized version of $ \theta $. The quantization
range is given by  $ \delta $. We assume that the density functions $ p(\theta) $ and $
p(\hat{\theta}) $ are smooth enough. So, we have $ P(\theta \in
[\hat{\theta}-\delta/2, \hat{\theta}+\delta/2]) \approx \delta p(\hat{\theta})
\approx  \delta p(\theta) $. The first approximation holds by the smoothness and the
second by the uniform distribution of $ \theta $ in the small enough range $
[\hat{\theta} -\delta/2, \hat{\theta}+\delta/2] $. Here, the operators should be
understood as element-wisely as $ \theta $ is a multidimensional variable.

Taking logarithm on the right side, we get the description length for both the
model (specified by parameters $ \theta $) and the observations under this
model. While it is impossible to evaluate its value with the true but unknown
parameters, we may approximate this quantity by the truncated ones. The total
code length is written as:
\begin{align}
\begin{split}
\length(\hat{\theta},Y) & \approx -\log p(\theta) - \log p(Y|\theta) \\
& \approx -\log(\delta p(\hat{\theta})) - \log p(Y|\hat{\theta})
\end{split}
\end{align}

The particular form of MDL adopted in our study is derived in Appendix, and our derivation
leads to the following criterion:
\begin{align}
\label{eq:min:obj}
\begin{split}
\{\hat{\theta}, d\} & = \argmin_{\theta,d} \big \{ -p(\theta) - \log p(Y|\hat{X},\theta) \\
& + \frac{1}{2} \log \det F_{X,\theta} F_X + \frac{d}{2}(1 + \log \kappa_d - \log 2 \pi) \big \}
\end{split}
\end{align}
where the minimization should be understood as simultaneously on model
parameters $ \theta $ and model order $ d $. Constant $ \kappa_d $ relates to
the geometric property of a $ d $-dimensional space. $ F_{X,\theta} $ and $
F_{\theta} $ are two terms relates to the curvature or second order expansion of
likelihood function.

By approximating $ F_{\theta,X} $ and upper bounding $ F_\theta $, we obtain the
quantity reflecting the code length that would be necessary for describing 
an LDS.

\begin{align}
\label{eq:obj}
\begin{split}
\{\hat{\theta}, d\}  &= \argmin_{\theta,d} \big \{ -p(\theta) - \log p(Y|\hat{X},\theta)  \\
&+\frac{1}{2} \log \det [C Q C^T +R_1]  + \frac{d}{2} \log \frac{2 N^2}{(2 \pi)^2}  \big \}
\end{split}
\end{align}
where $ Q $ is a positive definite matrix and is the solution to the discrete time Lyapunov equation (see, for example \cite{Khalil2001,Torokhti1961}) of the LDS. 

In particular, we can see that the description length correlates the system
stability through the a positive definite matrix $ Q $, which is closely related
to the stability of a dynamic system. When Lyapunov equation has no positive
definite matrix as its solution, a system is said to be unstable. In that case,
a minor deviation on the parameters from the true ones might be amplified
through the model and results in large deviation on system trajectory.
Therefore, we need more precisions to describe the parameters such that the
system deviation stays within a reasonable range. The converse also holds, when
the system is (asymptotically) stable and its system behavior is less sensitive
to parametric deviations. In that occasion, a coarse precision would be enough,
which means a small amount of description length. Therefore, from our analysis,
we know that the stability of LDS and description length are not totally
separable.


The proposed criterion could be used in many ways. Direct usage on a set of
candidate models is also possible. Herein, we make use of the criterion of
minimum description length for latent dimension annihilation. Specifically, let
$ d $ be an arbitrary model order and under this premise, we reduce the model
order by letting some of the latent dimensions be zeros. This philosophy does
not adopt the common two-stage inference procedure but directly aims at
identifying the most appropriate model order starting from the most complex one.

However, due to the unclear coupling between model selection and model
inference, the redundant latent dimensions are indistinguishable. Therefore, we
adopt the strategy of reinitialization as is standard in the EM algorithm when
we change the latent dimension. In addition, the necessary condition for which
an LDS being stable is also enforced to hold in the implementation. 

\begin{algorithm}[t]
	\caption{The complete algorithm for latent dimension annihilation.}
	\label{alg:main}
\begin{algorithmic}[1]
	\STATE \textbf{Inputs:} $ d_{min} $, $ d_{max} $, the initial guesses for parameters $ \theta = \{d,W,V,R_1,R_2\} $
	\STATE \textbf{Output:} The model parameters $ \hat{\theta} $, containing the latent dimension $ d $
	\STATE \textbf{Init:} $ t\leftarrow 0 $, $ d_{cur} \leftarrow d_{max} $, $ \mathcal{L}_{min} \leftarrow \infty $, $ dl^\star \leftarrow \infty $
	\WHILE {$ d_{cur} > d_{min} $}
	\REPEAT
	\STATE $ t \leftarrow t+1 $
	\STATE $ Z_t \leftarrow \E [\bm{x}_t \bm{x}_t|Y] $, $ Z_{t,t-1} \leftarrow \E[\bm{x}_t \bm{x}_{t-1}^T|Y] $
	\STATE $ R_1^{new} \leftarrow \dfrac{1}{T-1} (\sum Z_{t} - A^{new} \sum Z_{t-1,t}) $
	\STATE $ R_2^{new} \leftarrow \dfrac{1}{T}\sum(y_t) $
	\STATE $ \pi^{new} \leftarrow \bar{x}_1 $
	\STATE $ C^{new} \leftarrow (\sum y_t \bar{x}_t)(\sum Z_t)^{-1} $
	\STATE $ A^{new} \leftarrow (\sum Z_{t,t-1})(\sum Z_{t-1})^{-1} $
	\IF {$ \max\{\operatorname{eig}(A^{new}) \} > 1 $}
	\STATE $ A^{new} \leftarrow A^{new}/\operatorname{diag}(1.1*\max(\operatorname{eig}(A^{new})) $
	\ENDIF
	\STATE $ \theta(t) \leftarrow \{A^{new}, C^{new}, R_1^{new},R_2^{new}, \pi^{new},d_{cur} \} $
	\UNTIL {$ |\mathcal{L}(\theta(t-1),Y) - \mathcal{L}(\theta(t),Y| < \epsilon $}
	\IF {$ \mathcal{L}(\theta(t),Y) \le \mathcal{L}_{min}$}
	\STATE $ \mathcal{L}_{min} \leftarrow \mathcal{L}(\theta(t),Y) $
	\STATE $ dl(d_{cur}) \leftarrow \length(\theta(t),Y) $
	\ENDIF
	\IF {$ dl(d_{cur}) <  dl^\star$}
	\STATE $ dl^\star \leftarrow dl(d_{cur}) $, $\theta^\star \leftarrow \theta(d_{cur})$, $ d^\star \leftarrow d_{cur} $
	\ENDIF
	\IF {$ dl(d_{cur}) > dl(d_{cur} - 1) $}
	\STATE \textbf{break}
	\ENDIF
	\STATE $ d_{cur} \leftarrow d_{cur} - 1 $
	\ENDWHILE
\end{algorithmic}
\end{algorithm}

\begin{figure*}[ht]
	\begin{center}
		\subfloat[LDS of latent dimension
		eight.]{\includegraphics[width=0.45\linewidth]{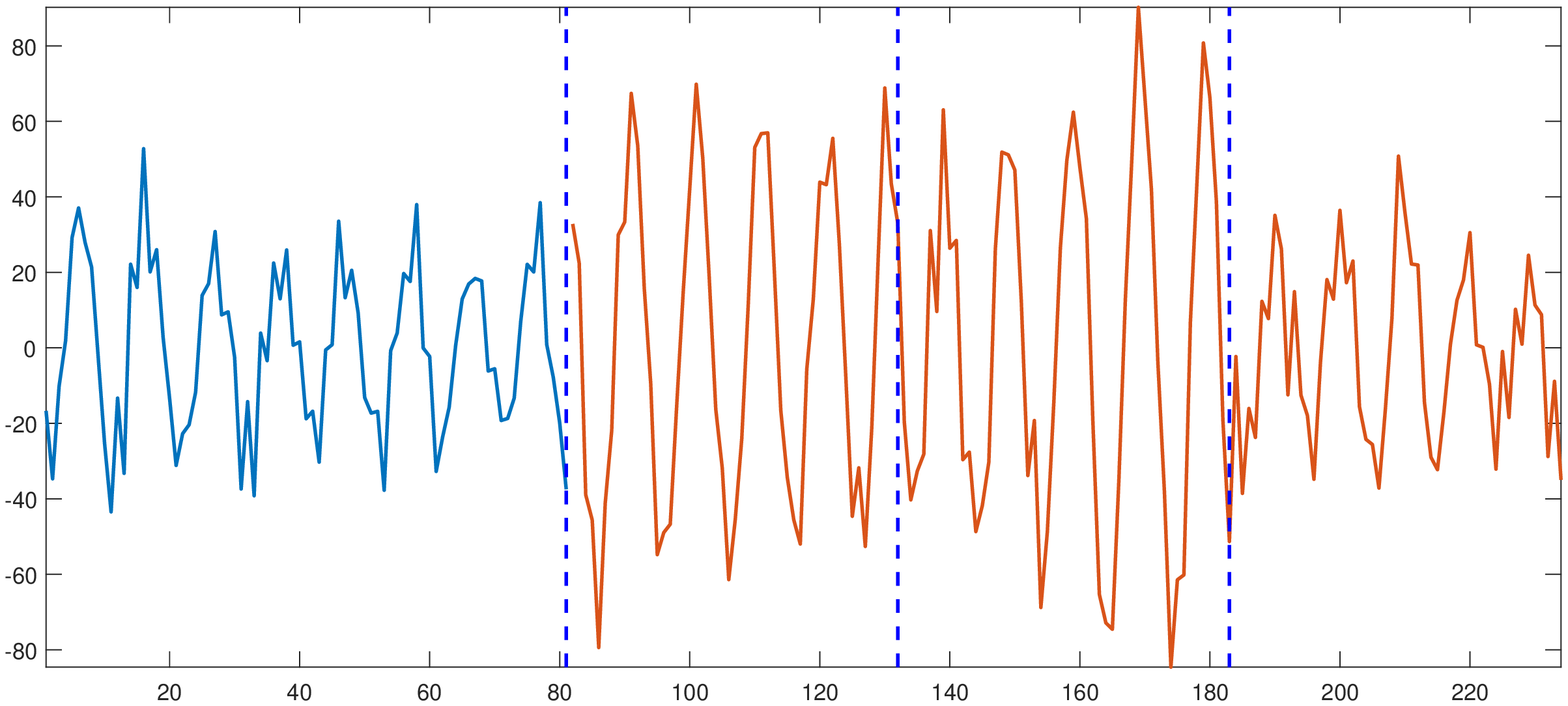}} \hfil
		\subfloat[LDS of latent dimension
		six.]{\includegraphics[width=0.45\linewidth]{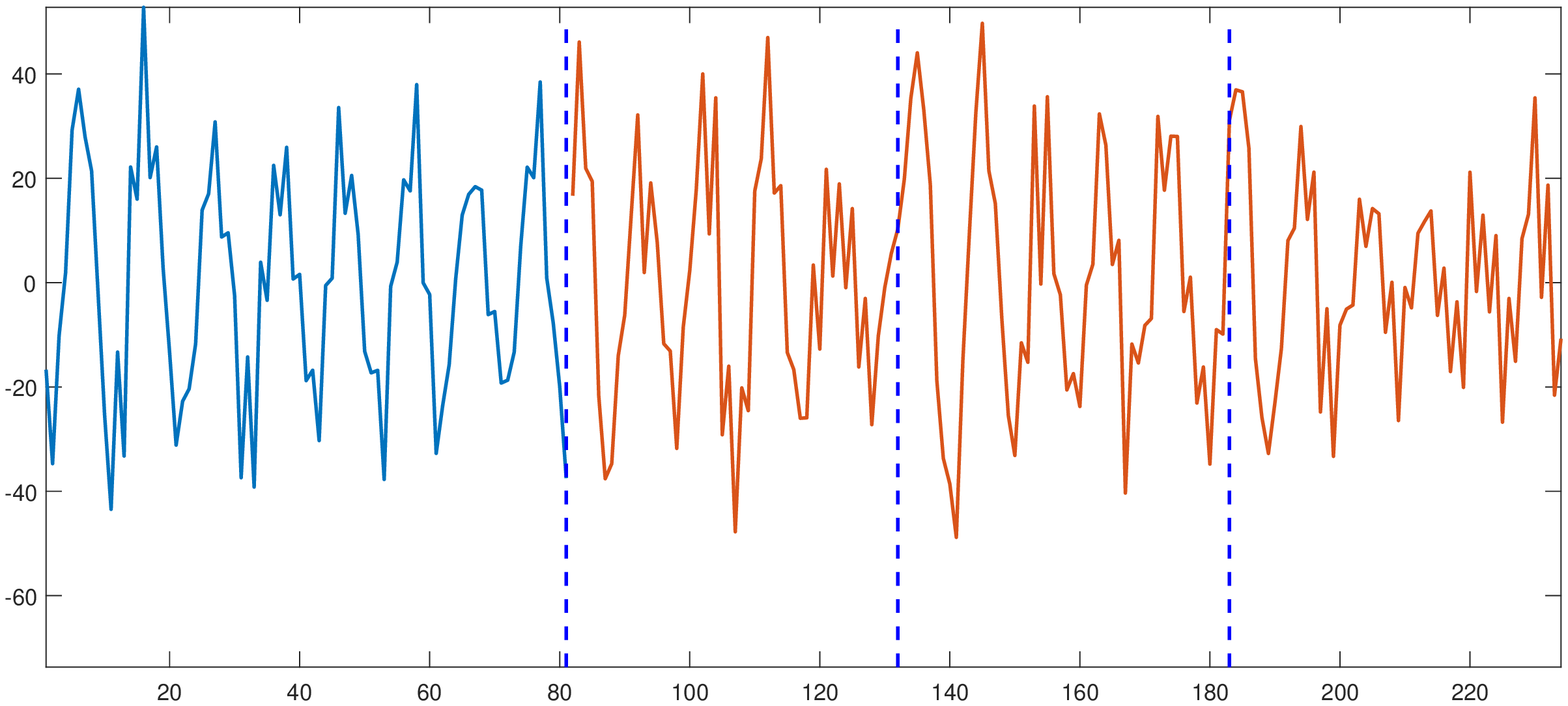}}\\
		\subfloat[LDS of latent dimension
		four.]{\includegraphics[width=0.45\linewidth]{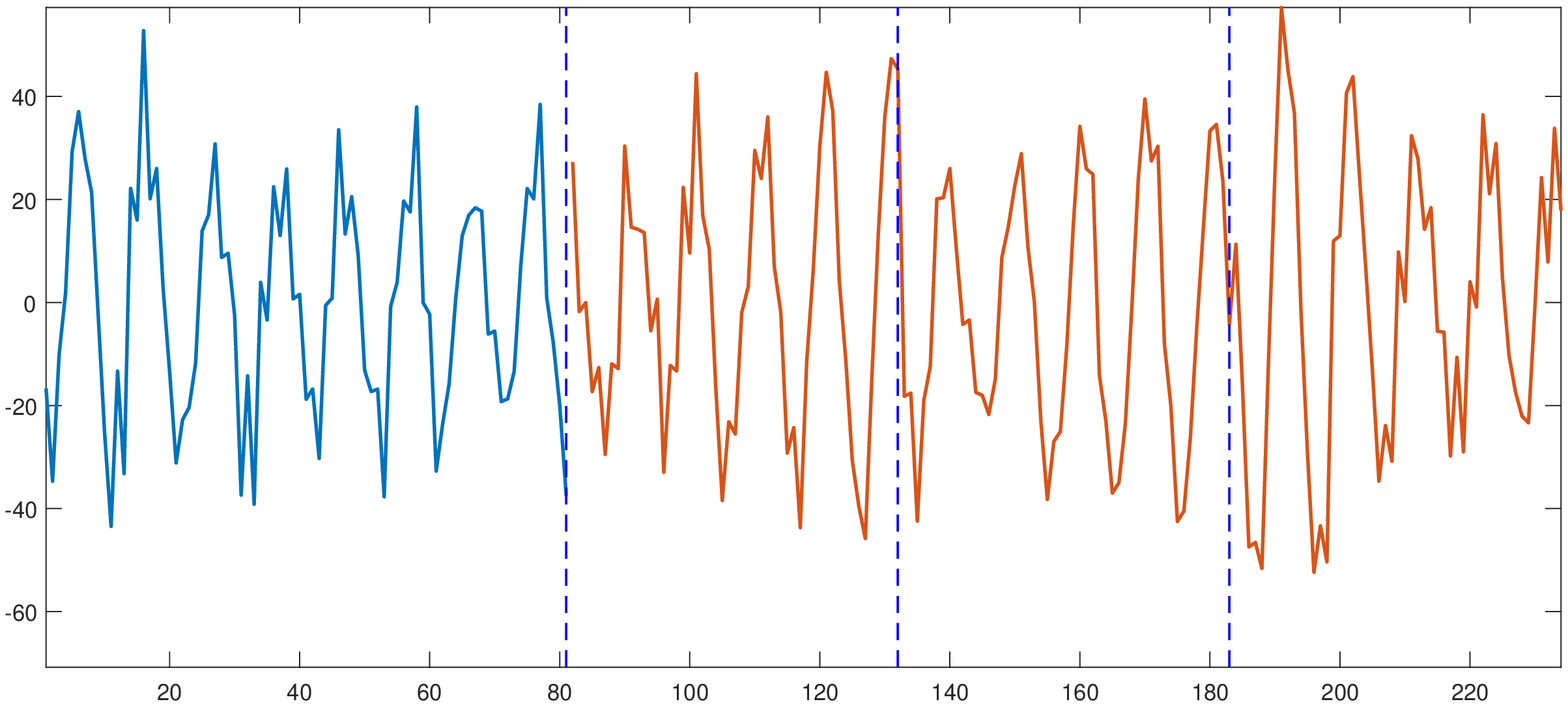}} \hfil
		\subfloat[LDS of latent dimension
		two.]{\includegraphics[width=0.45\linewidth]{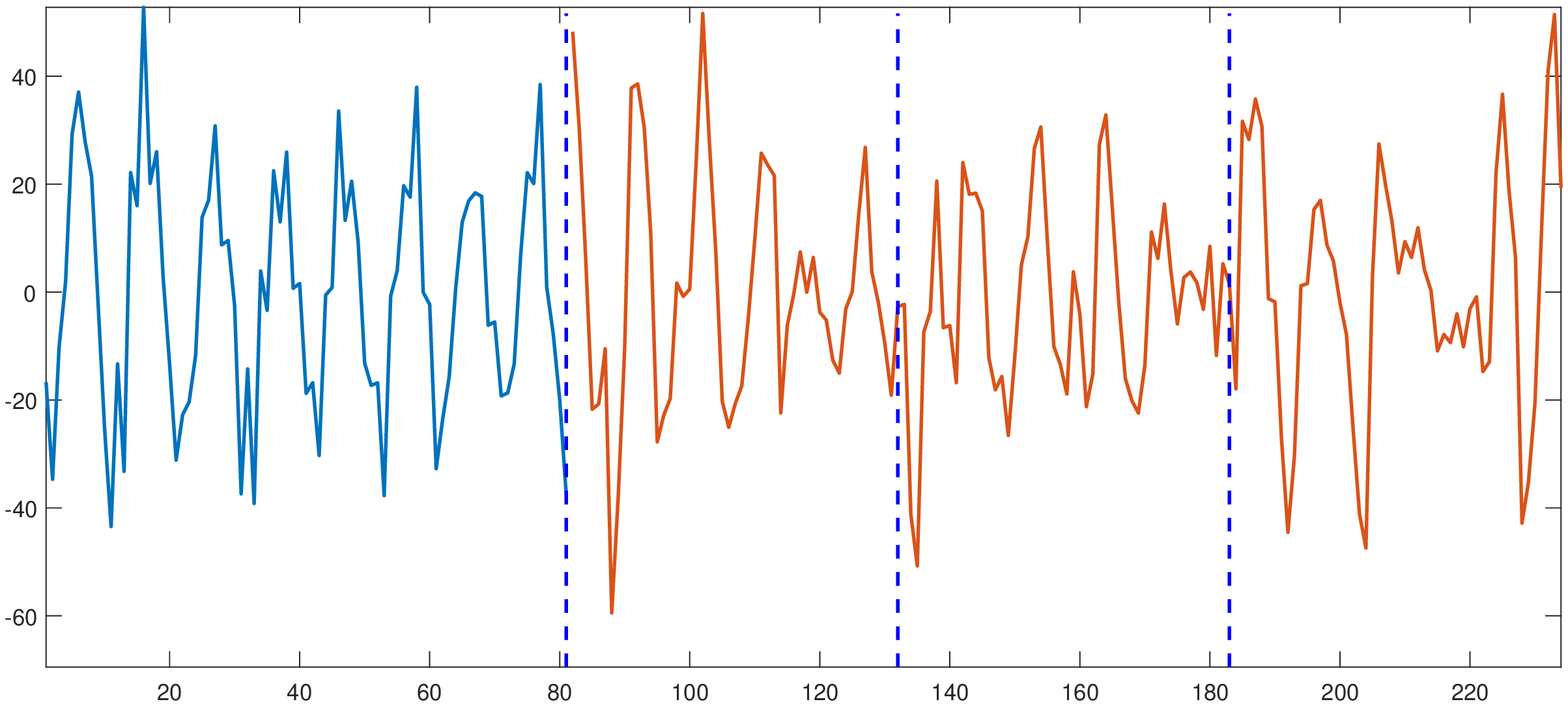}} \caption{The
			comparison of generated sequences of different model orders. The training
			sequence in the first part of figures was generated from an LDS of order four.
			Within all four sample sequences, the model of order four best regenerate
			similar observations as the training sequence. The fitted model of this order
			is also the minimum of a list of candidate models in terms of description
			length. This result coincides with the newly proposed principle MDL.
			\label{fig:lds:varorder}}
	\end{center}
\end{figure*}

The complete algorithm is listed in \cref{alg:main}. The general procedure
follows the EM algorithm. For an LDS, E-step constitutes of computing the
expected log likelihood for complete data. The M-step infers parameters by
taking the partial derivatives of the log likelihood obtained in E-step with
regard to the unknown parameters and setting the results to zero. In line 7,
the expectation is taken over the entire data. It differs from the one computed
in the Kalman filter where only past observations are used. In line 13-15
\cref{alg:main}, the condition for which an LDS being stable is checked and if
this condition does not hold the eigenvalues of its transition matrix are
rescaled. The function $ \operatorname{diag}(\cdot) $ repeats and aligns its
scalar argument along the main diagonal. The $ \length(\cdot) $ computes the
description length that is used in comparisons among models. As we will
demonstrate in the experiments, this shift from two-stage model selection could
always lead to a consistent approach for a given set of observations.

\section{Experiment}
\label{sec:exp}

\subsection{Sequence prediction:}
In this experiment, we adopted a one-dimensional LDS to
generate synthetic observations. Its latent dimension was set to four. The
observatory matrix and transition matrix were both generated randomly from the range of
$ [-1,1] $. In addition, the transition matrix was preprocessed such that the
maximum eigenvalue was less than one on the magnitude. The covariance matrices in
noises terms were sampled from inverse Wishart distribution with scale matrix
being an identity matrix.  In Bayesian statistics, this distribution is usually adopted
as the conjugate prior for the covariance matrix in a multivariate normal
distribution. The sequence of observations was recorded after a short
\emph{burn-in} period (20 long). The observable sequences were 100 in
length. We set the maximum model order $ d_{max} $ for testing to twelve and the minimum model order 
to two. 

\begin{figure}
	\begin{center}
		\includegraphics[width=0.95\linewidth]{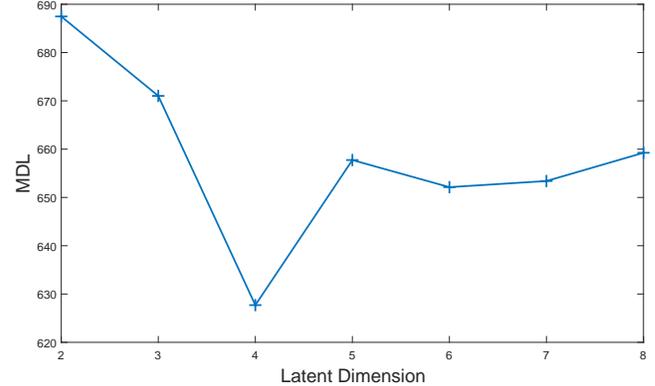}
		\caption{The description length of the synthetic data. The observations to be
			fit for were generated from an LDS with latent dimension being four. By the
			principle of MDL, a model is selected if it yields the shortest description
			length, as it best extracts the regularities from the observations. Thus the
			model with latent dimension four should be selected and this result coincided
			with the setting of data generator. More details on experimental configuration
			are given in \cref{sec:exp}. \label{fig:syn:1:mdl}}
	\end{center} 
\end{figure}

The ML estimates of parameters were adopted. Then the description
lengths were computed according to \cref{eq:min:obj}. Within the principle of MDL,
the one that leads to a minimum description length should be selected as it
provides the most parsimonious descriptor for data and is considered to generalize well
over unseen observations \cite{MyungJay2018}. The values of description lengths are depicted in
\cref{fig:syn:1:mdl}. In our experiment, the model of latent dimension four achieved the minimum
value. This result was in accordance with the data generator
as described previously.

\begin{figure*}
	\centering \subfloat {\includegraphics[width=0.45\linewidth]{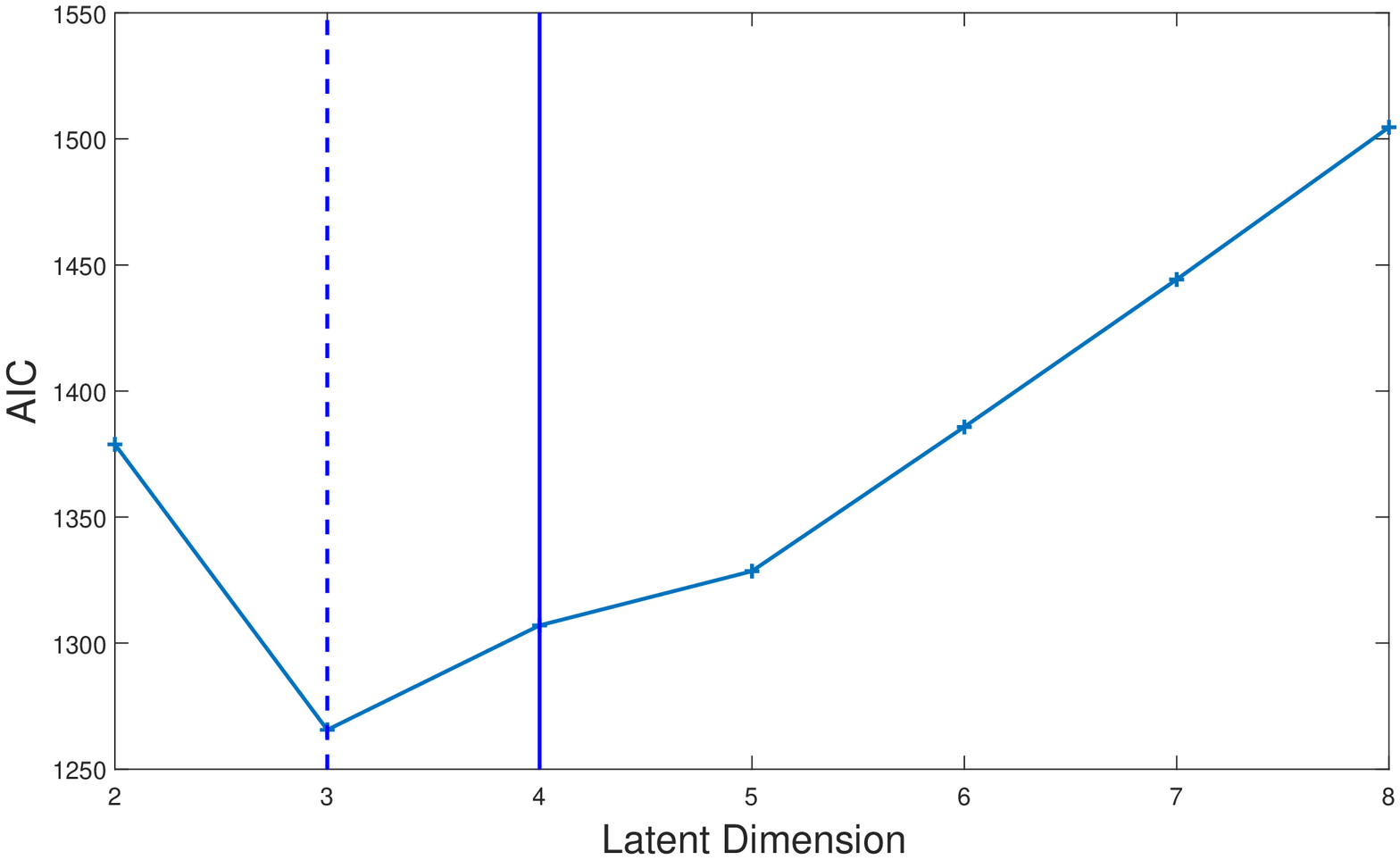}}
\hfil \subfloat {\includegraphics[width=0.45\linewidth]{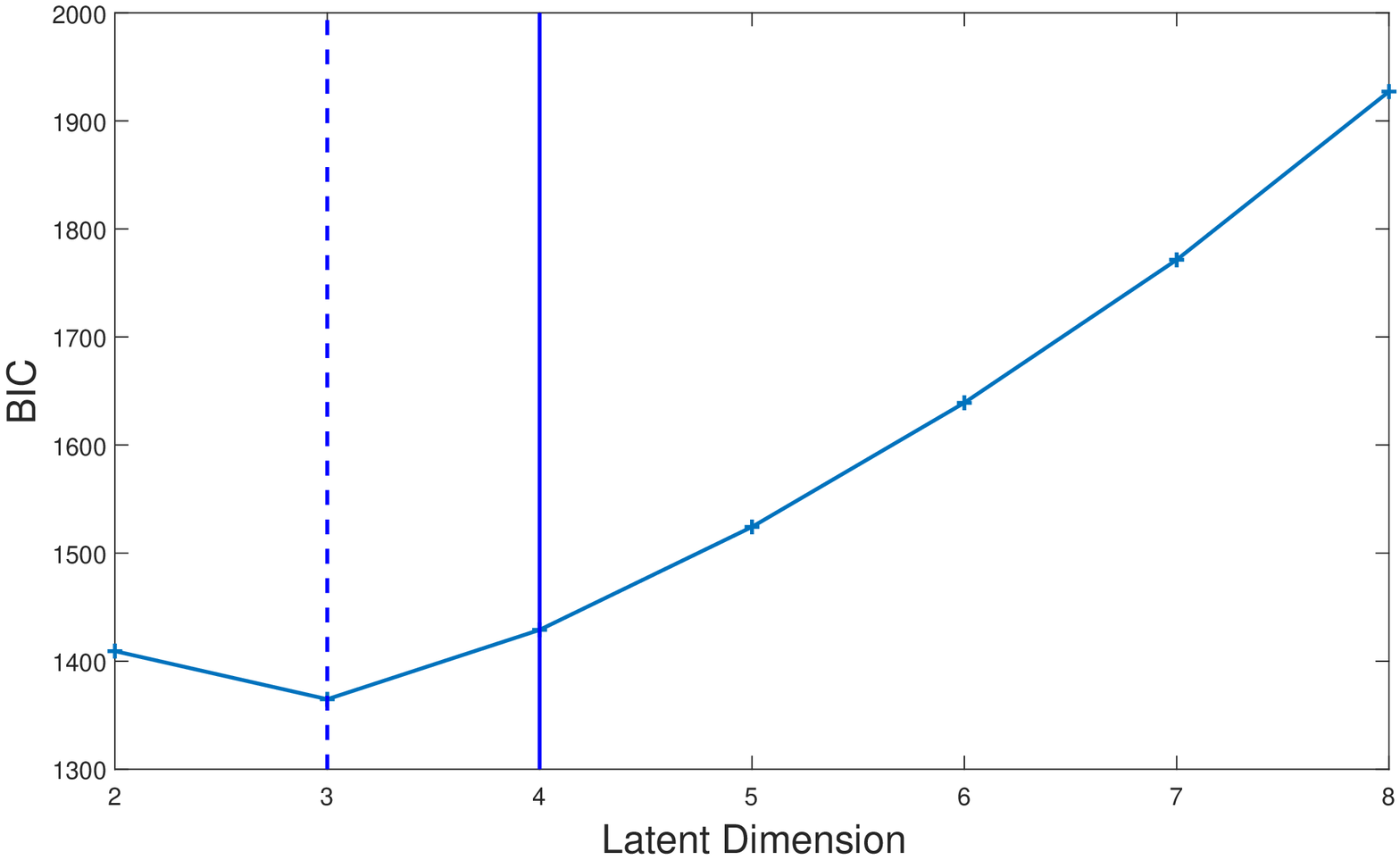}}
\\\vfil \subfloat {\includegraphics[width=0.45\linewidth]{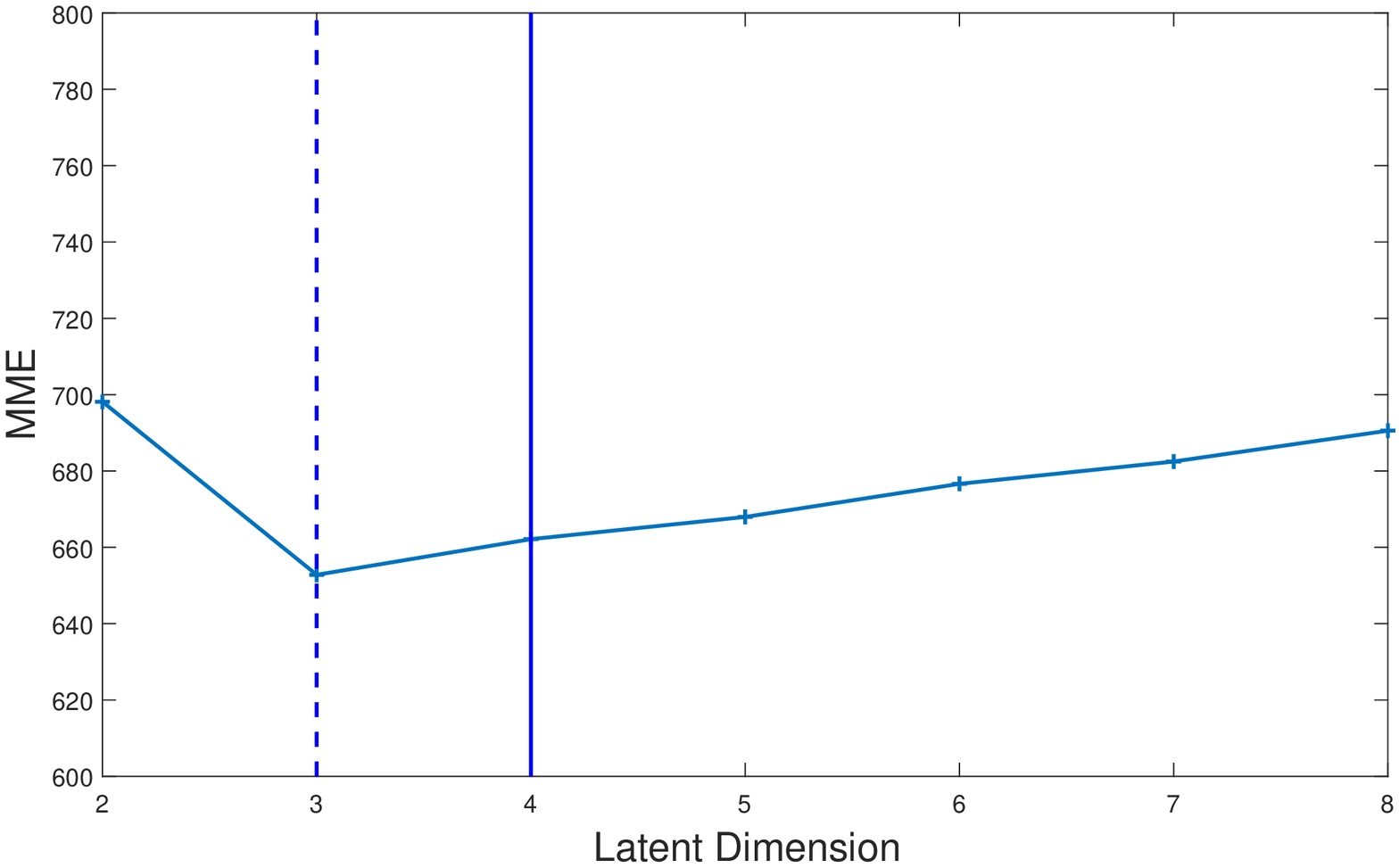}}
\hfil \subfloat {\includegraphics[width=0.45\linewidth]{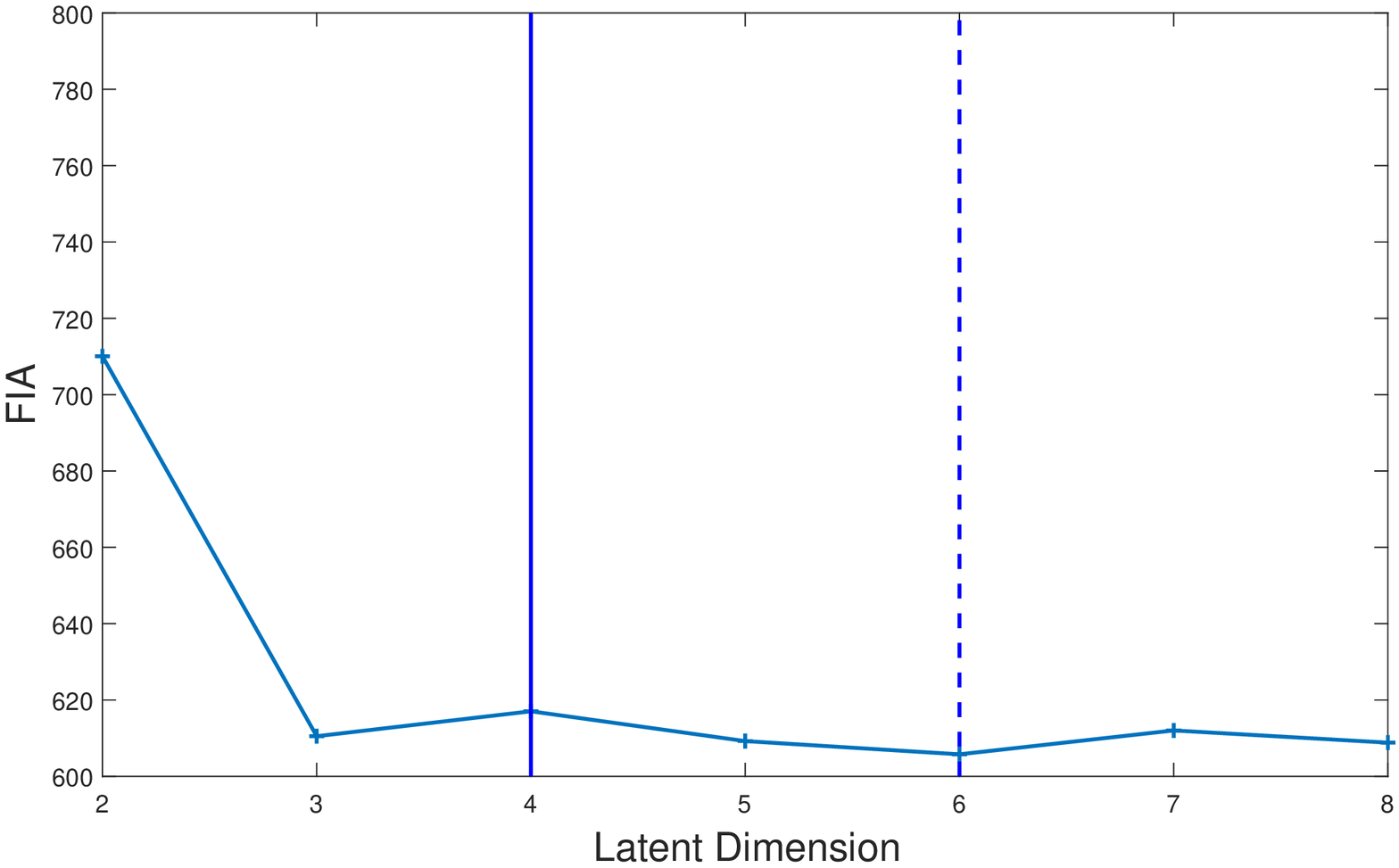}}
\caption{The values of various criteria in the task of model selection. The
	solid lines in figures indicate the true latent dimension, of which the model
	generated the training sequence. The dashed line indicates the dimension chosen by
	various criteria. In this example, AIC and BIC have similar behavioral
	properties and are prone to choose a model with less complexity. MME also
	erroneously converges to the latent dimension three. FIA inclines to choose 
	dimension six, yet it does not strongly support for this particular latent dimension.
	In fact, loosely speaking, it favors a list of dimensions where their description
	lengths are close on magnitude. \label{fig:var:cra} }
\end{figure*}

To compare the performances of fitted LDS's of various orders, we used the
fitted models to generate sample sequences of length 1000 and intercepted three
representative parts from the beginning (20\%, after burn-in period),
intermediate (50\%), and the end. Each segment was equally long. We concatenated
the sampled subsequences together and plotted them along with the original
synthetic data. For clarity of comparison, we only took samples from models of
every second order. The results are shown in \cref{fig:lds:varorder}. From the
figure, we observe that LDS with latent dimension four is clearly the one that
is able to generate a similar sequence as the original one. More importantly,
the sequences generated by this model could carry the resemblance till the end.
Other three models could also generate a similar sequence in the beginning, and two of them
could maintain this resemblance till the middle part, but few could keep it till
the end.

Apart from MDL, we also tested different criteria on this artificial data. The
comparison criteria include: (1) Akaike information criterion, AIC; (2) Bayesian
information criterion, BIC; (3) Fisher information criterion, FIA; and (4)
Minimum message estimate, MME \cite{Figueiredo2002}. Specifically, FIA and MME
are also built on the same ground of description length as ours. The MME is
proposed especially for a finite Gaussian mixture model. FIA approximates the
model complexity by a term relevant to the model geometry, which has been shown
to be closely related to the model capability. The results are depicted in
\cref{fig:var:cra}.  From the experimental results, AIC, BIC, and MME selected a
model of a smaller order than the true one. In contrast, the values of FIA on a range
of latent dimensions were close and no particular dimension showed clear
advantage.

\subsection{Real-world application}

In this experiment, we consider a real-world data set, \emph{Character}, from
UCI machine learning repository \cite{Dheeru2017}. This data set consists of character samples
which have been differentiated and Gaussian smoothed beforehand. We used the
Euclidean trajectories of pen tip from the data.

\begin{figure}
	\centering
	\includegraphics[width=0.9\linewidth]{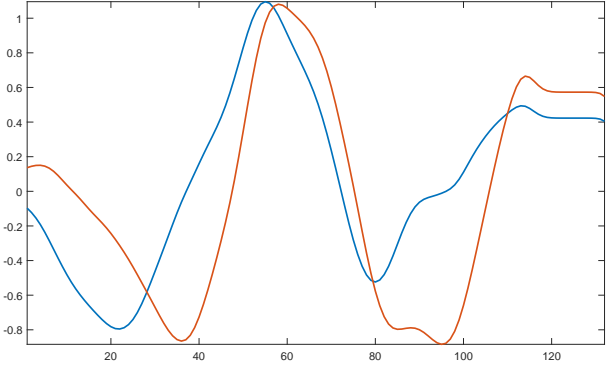}
	\caption{Illustration of Euclidean trajectories of the pen tip. This example was taken from \emph{Character}. The data have been numerically differentiated and smoothed.}
	\label{fig:real-char-demo-1-mdl}
\end{figure}

An example from the data is given in \cref{fig:real-char-demo-1-mdl}. Other
samples in this data set have almost the same smooth trajectory. The
smoothness and irregular curvature in data propose challenges to the model selection.
Higher order models are more prone to overfit the sample sequence and thus
generate too flexible curves. Meanwhile, lower order models are less likely to
overfit but easier to produce rigid and non-smooth sequences.

For the convenience of external comparison, we normalized the quantities
computed in terms of different criteria and placed them together. We
took the example depicted in \cref{fig:real-char-demo-1-mdl} as the input
and computed the quantities with regard to various criteria. To remove the
randomness brought by initialization in EM, we adopted the strategy of
reinitialization in EM and picked out the one with the least squared prediction loss. 

\begin{figure}
	\centering \includegraphics[width=0.9\linewidth]{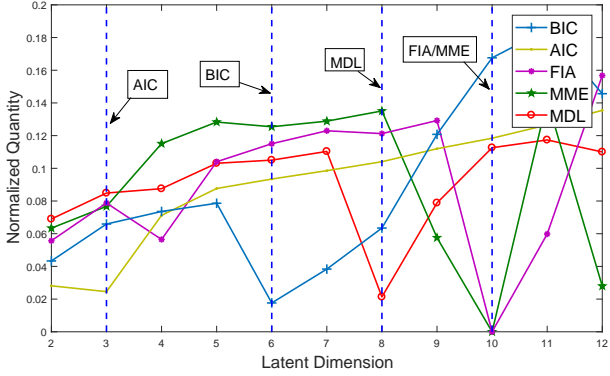}
\caption{The quantities computed with regard to various criteria. The chosen
	orders are marked by a dashed line and indicated by an annotation. The maximum
	latent dimension was set to twelve, $ d_{max} = 12 $, and the minimum order was
	set at two, $ d_{min} =2 $. For the convenience of external comparison, the
	quantities were normalized. } \label{fig:real-char-1-norm-aug}
\end{figure}

The quantities computed according to different criteria are depicted in
\cref{fig:real-char-1-norm-aug}. From the figure, it can be inferred that, due
to a lack of consideration on structural properties of the models, AIC and BIC
opted for a model with less complexity or of smaller order. In our current
experimental setting, FIA and MME had a similar preference and both chose the model
of latent dimension ten as the ``optimal'' model in terms of the chosen
criteria. MDL, on the other hand, selected an intermediate latent dimension between the
ones chosen by BIC and FIA/MME.

\begin{figure}
	\centering
	\includegraphics[width=0.9\linewidth]{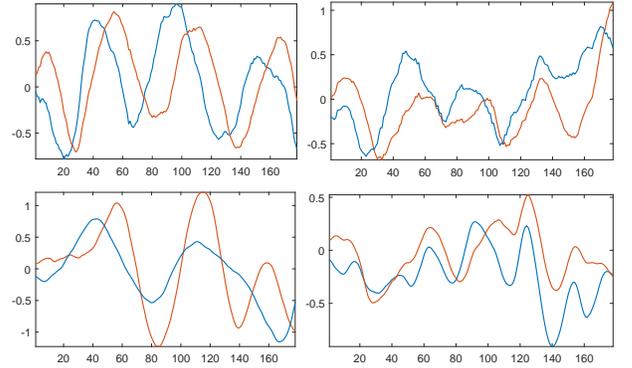}
	\caption{Demonstration of trajectories generated by fitted models. The subplots are for models of every second order from six to ten. They are aligned clockwise in increasing order. The sequence to be fitted is shown in \cref{fig:real-char-demo-1-mdl}. Based on the generated sample sequences, the model of order eight is more comparatively suitable than other three candidate models.  }
	\label{fig:real-char-aug-demo}
\end{figure}

We plotted the approximated trajectories yielded by selected ``optimal'' models
in terms of different criteria in \cref{fig:real-char-aug-demo}. From the
figure, it can be seen that models with small latent dimensions were not
flexible enough for replicating the patterns exhibited by the input. Take the model
with latent dimension two as an example (left-up panel in
\cref{fig:real-char-aug-demo}). The yielded sample sequences exhibited a regular
pattern, which only matched the input sequence coarsely. Meanwhile, the curves
were corrupted by noises. Conversely, the model with a higher order (right-below
panel in \cref{fig:real-char-aug-demo}) could generate smooth curves yet run
the risk of dedicating much model capability to fitting for the irregular pattern in
the input. This could be seen in the last panel where the model
produced curves with severe oscillation.

\subsection{Latent variable annihilation}

\begin{figure}
	\centering \includegraphics[width=0.9\linewidth]{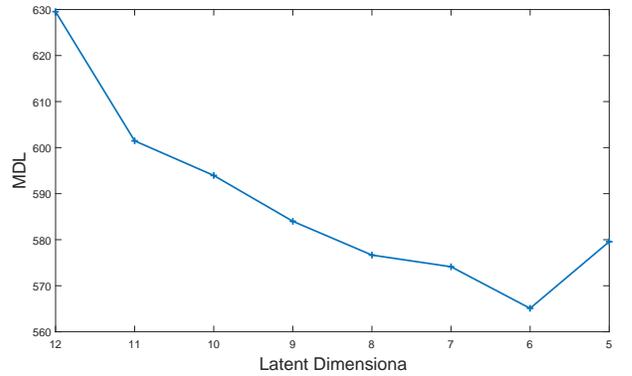}
	\caption{The changes on description in a range of model orders. For
		latent variable annihilation, we started with the highest-order model class,
		and then reduced its latent dimension by one in each step. \label{fig:devar-demo-mdl-1}}
\end{figure}

\begin{figure*}
	\centering
	\includegraphics[width=0.85 \linewidth, height=1.5in]{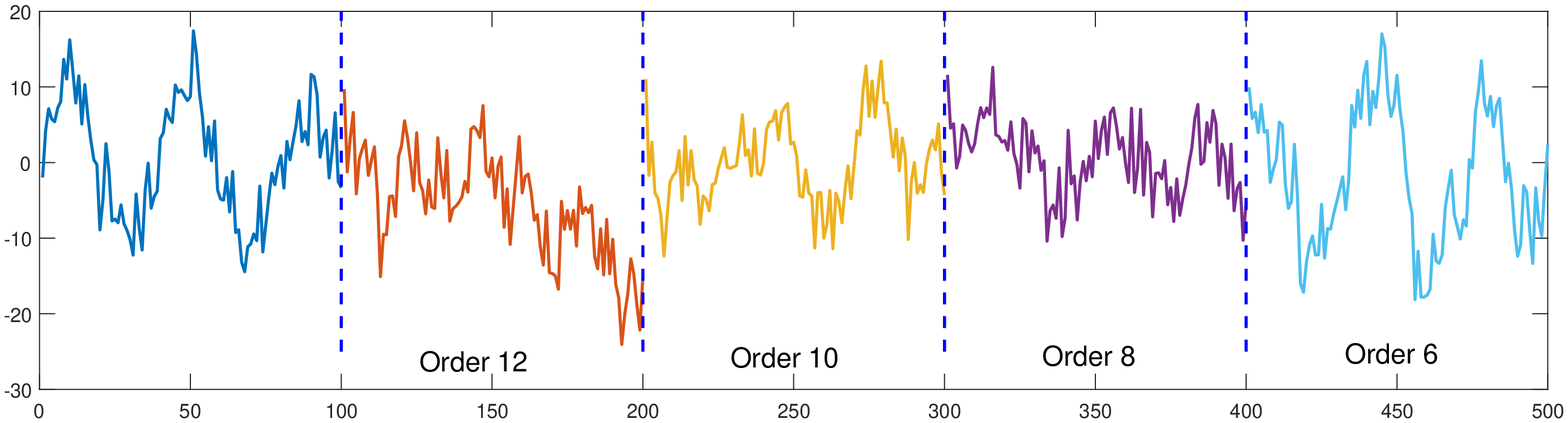}
	\caption{The training data and sample segments from models of different orders. The training data was displayed in the first panel. In other panels, the sample segments were shown. The samples were all intercepted after an initial burn-in period, which was to reduce the randomness brought by the initial state. The model of order six was the one having a minimal description length. Its sample segment was shown in last panel. \label{fig:devar-demo-seq}}
\end{figure*}

In this experiment, we adopted an LDS with latent dimension being six. As before, the observatory matrix and transition
matrix were generated randomly in the real interval $ [-1,1] $. The transition
matrix was further processed such that the yielded LDS satisfied the stability
condition. The covariance matrices of noise terms were also generated according
to the inverse Wishart distribution with an identity scale matrix. In our
experimental setting, the maximum model order was set to twelve. We started from
this model order $ \mathcal{M}_{d_{max}} $, then reducing its complexity by
forcing one latent variable to be zero each time. The model was retrained and
reassessed in terms of description length. We stopped this procedure when the
criterion value no longer decreased. According to the philosophy of MDL, the one
which obtained the minimum value on the chosen criterion was
considered the ``best'' model.

We reported in \cref{fig:devar-demo-mdl-1} the changes in terms of description
length. The model with order six was the one that achieved the minimum value.
According to the MDL, this model order was the desired one. A trade-off between
model complexity and model fitness was achieved. A model with a higher
complexity came with a higher description length, in which the penalty on model
complexity contributed more to the summand in the formula of description length.
When the model became less complex, the loss on goodness-of-fit was complemented
by the decrease of the penalty on the model complexity. In the order of six, the
sum of goodness-of-fit and penalty was the lowest. For models of smaller orders,
the loss of goodness-of-fit became severe and thus resulted in an increase of
the description length.

We included in \cref{fig:devar-demo-seq} a list of sample sequences generated
from fitted models. The first segment was the training data. From left to right,
the segments were generated from fitted models in decreasing model order. In
particular, the first one was the one that with the highest order, and the last
one was the one chosen by the principle of MDL. The two segments in between were
from models of order ten and eight respectively. In this example, we can see
that a complex LDS easily dedicated its model capacity in fitting for the noises
and thus not ideally replicated the patterns shown in data. This phenomenon
became less severe when model order turn small. Eventually, the model of order
six generated a the segment that resembled training
data most.

We investigate the effect of data amount on the model selection using a
real-world data set. In this experiment, we adopted \emph{gesture} from UCI
machine learning repository \cite{Dheeru2017}. This data set consists of
readings of positions of hands, wrists, head, and spine of a user in each video
frame, as well as velocity and acceleration of hands and wrists. More
description of this data set is given in \cite{Madeo2013}. This data set was
collected for classification, and each sequence corresponds to a label. We used
the sequences in the first class and sampled from them five batches. The first batch made up 20\% percent of the entire sequences in this class, and the second batch made up 40\%, and so forth. This step
made sure that the data used as input were homogeneous. That is, they should all
be explained by a same model. Each batch was referred by attaching its
percentage (without \%) to the data name. 

The maximum latent dimension was set to ten, $ d_{max} = 10 $ and minimum latent
dimension was set to two, $ d_{min} = 2 $. Other configurations kept the same
with the real-world data experiment in previous section. The results were reported in \cref{tab:1}.
As there was no ground-truth model order for this real-world data, and each
criterion may offer different suggestions, the dimension chosen as reference is
the one that minimizing the criterion value under the whole data in the class. In the table, the reference
dimensions were placed along with the data batches. The value pair, i.e. the criterion value as well as the
normalized one, for different percentages were collected and reported in
\cref{tab:1}. When an inference criterion made a choice differing from the
reference dimension, it is marked as italic and by a superscript asterisk, the
chosen dimension was labeled in the subscript. 

\begin{table*}
	\caption{The quantities computed according to various criteria on different amounts of data. More description is given in the text. \label{tab:1}}
	\centering
	\begin{normalsize}
		\begin{tabular}{|c|l|l|l |l|c|}
			\hline 
			& \emph{gesture\_20} & \emph{gesture\_40} & \emph{gesture\_60} & \emph{gesture\_80} & \emph{gesture\_100}  \\ 
			\hline 
			AIC (dim 2) & 0.04 (145.03) & 0.06 (252.35) & 0.04 (317.80) & 0.07 (503.50)  & 0.09 (611.40) \\ 
			\hline 
			BIC (dim 2) & 0.03 (154.20) & 0.04 (277.02)  & 0.04 (359.53) & 0.06 (514.52) & 0.08 (614.84) \\ 
			\hline 
			FIA (dim 7) & 0.05  (70.37) & 0.08 (85.82) & \textit{0.09 (183.25)}$ ^\ast_6 $  & \textit{0.12 (253.60)}$ ^\ast_6 $ & 0.08 (154.07) \\ 
			\hline 
			MME (dim 8) &  \textit{0.14 (115.88)}$ ^\ast_5 $ & 0.13 (214.36) &\textit{0.12 (231.55)}$ ^\ast_7 $ &\textit{0.12 (283.67)}$ ^\ast_9 $  & 0.08 (228.65) \\ 
			\hline 
			MDL (dim 7) & \textit{0.10 (110.18)}$ ^\ast_6 $ & 0.10 (164.44) & 0.10 (179.04) & 0.09 (221.13) & 0.08 (272.41) \\ 
			\hline
		\end{tabular}
	\end{normalsize}
\end{table*}

From the table, it can be seen that AIC and BIC tended to recommend the least
complex model in the candidate model set. Their results would not change with the
increase of training data. Meanwhile, FIA, MME, and MDL made more reasonable
recommendations compared with AIC and BIC. Within all three criteria, in terms
of consistency on model selection, MDL was superior than other two criteria.
There was only one different recommendation when it was applied on the first twenty
percent data. FIA and MME made respectively two and three different selections with regard to
their reference dimensions. 

\begin{figure}
	\centering
	\includegraphics[width=0.9\linewidth]{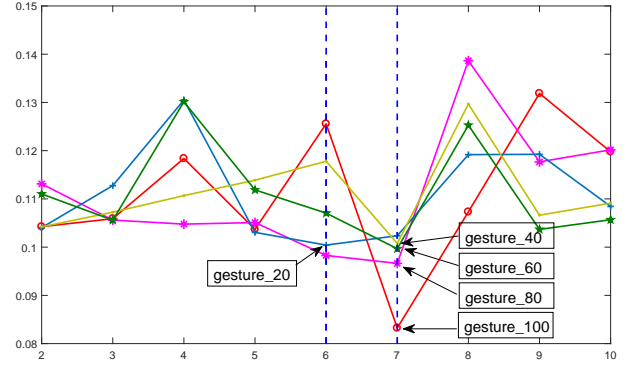}
	\caption{The interior comparison of MDL when training on different amounts of data. We can see that when the amount of input is increasing, the dimension shows more relative advantage in the interior comparison. By the principle of MDL, this dimension gets better trade-off between goodness-of-fit and model complexity as training data increase.  \label{fig:mdl:dataaug} }
\end{figure}

Another phenomenon that worth noting is that, with the increase of data amount,
the normalized value of description length calculated on the reference dimension
(seven in our example) gets smaller, which means that, under the criterion of
MDL, this dimension gains more relative advantage within the interior
comparisons. The increase of data amount makes the goodness of balance between
model complexity and model fitness more obvious in the selected dimension. This
result is more evident in \cref{fig:mdl:dataaug}. In the figure, when the amount
of data increases, the description length in the reference dimension gets lower
and lower. Therefore, by the principle of MDL, this dimension is perceived
as getting better trade-off between goodness-of-fit and model complexity as the
accumulation of data.

\subsection{Nonlinear synthetic data}

\begin{figure}[h]
	\centering
	\includegraphics[width=0.9\linewidth]{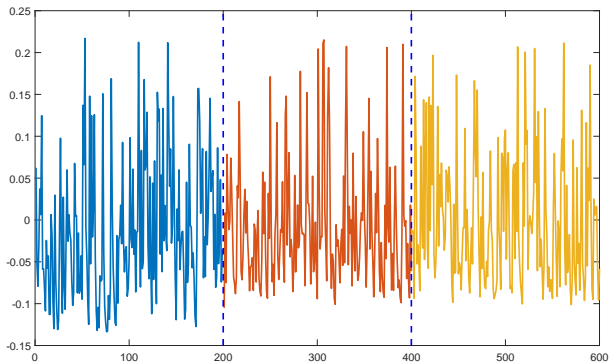}
	\caption{The synthetic sequences generated from models with different orders. }
	\label{fig:changepoint-det-demo-src-data}
\end{figure}

In this experiment, we tested the performance of MDL using sequences yielded
from nonlinear autoregressive–moving-average (NARMA) models. To do so, we let
the observability matrix be identity. Thus, the latent variables became
observable. The modified criterion was used to determine the suitable model
order for the NARMA sequences.

The three NARMA models are of order 10, 20, and 30. Their mathematical forms are given by:
\begin{align}
\begin{split}
x(t+1) &= 0.3x(t) + 0.05x(t)\sum_{i=0}^9x(t-i) \\
\qquad &+ 1.5u(t-9)u(t) + 0.1 \\
x(t+1) &= \tanh \big \{0.3x(t) +0.05x(t)\sum_{i=0}^{19} x(t-i) \\
&\qquad +1.5u(t-19)u(t)+0.01 \big \}+0.2 \\
x(t+1) &= 0.2x(t) + 0.004x(t)\sum_{i=0}^{29}x(t-i) \\
& \qquad + 1.5u(t-29)u(t) + 0.201
\end{split}
\end{align}
where $ u(t) $ were generated uniformly from interval $ [0,0.5] $. They
formed an \emph{i.i.d.} input stream for three models. We used each model to generate
one long NARMA sequence of length 1000. Each sequence was further processed by
subtracting its mean and removing outliers outside the interval $ [-0.5,0.5] $.
This step was to remove possible salient features from sequences. Some segments
from the generated sequences are illustrated in
\cref{fig:changepoint-det-demo-src-data}. To model these nonlinear data is challenging for MDL due to
the nonlinearity and the long-time correlations.

\begin{figure}[h]
	\centering
	\includegraphics[width=0.9\linewidth]{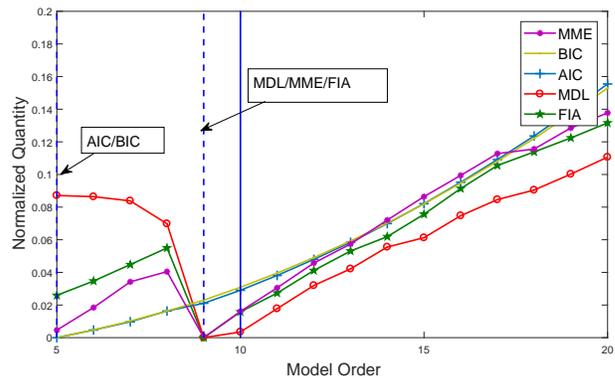}
	\caption{The performances of different criteria on sequence generated by the model of order 10. }
	\label{fig:aug-sys10}
\end{figure}

\begin{figure}[h]
	\centering
	\includegraphics[width=0.9\linewidth]{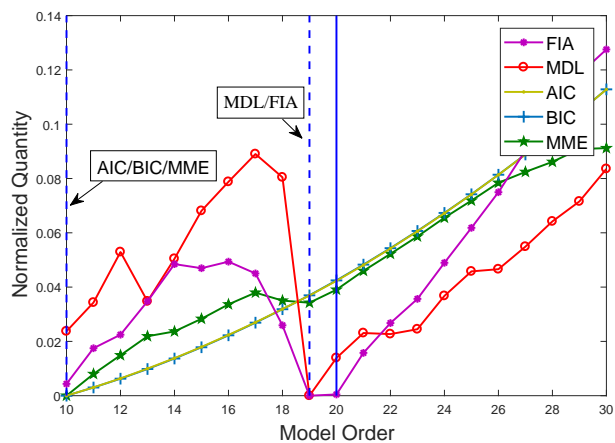}
	\caption{The performances of different criteria on sequence generated by the model of order 20. }
	\label{fig:aug-sys20}
\end{figure}

\begin{figure}[t]
	\centering
	\includegraphics[width=0.9\linewidth]{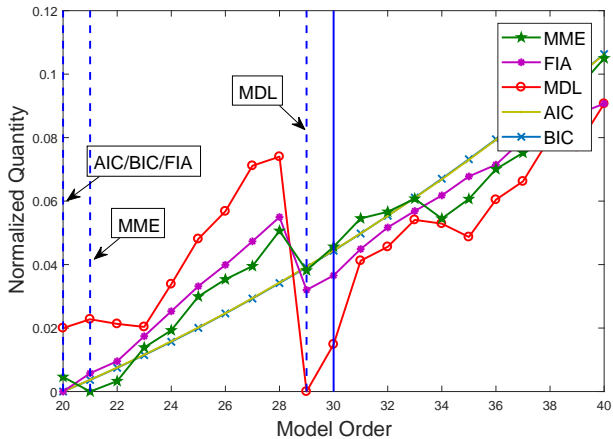}
	\caption{The performances of different criteria on sequence generated by the model of order 30. }
	\label{fig:aug-sys30}
\end{figure}

Since various criteria on model selection have different scales and the comparison is
carried out only within each criterion. Based on these two reasons, we
normalized all values in each criterion to facilitate the exterior comparison. The quantities computed in terms of each criterion are depicted in
\cref{fig:aug-sys10,fig:aug-sys20,fig:aug-sys30}. In this occlusion where underlying models were nonlinear, all the five criteria tended to choose a model that
was of less order than the true one. In all occasions, our proposed method, as
well as MME outperformed others in detecting the true model order. This suggests the
goodness of model-dependent penalty that considers the structural properties and
model capability. 

From \cref{fig:aug-sys10}, we can see that for large sample size, AIC and BIC
were both dominated by their respect penalty terms. As a consequence, they both favored a model that
was with the least complexity. In contrast, FIA, MDL, and MME got more reasonable
results as they all considered both the model capacity and the statistical
properties in devising their penalty terms. In fact, on order ten and nine, our proposed criterion reported similarly low quantities. 

For high model order twenty, it can be seen from \cref{fig:aug-sys20}, AIC, and BIC had a
similar performance as in the previous example. FIA, however, failed in
selecting an appropriate model order. FIA reported the same result as ours yet
its value on the true model order was also very low.

In the last occasion where exists long-time correlation among the sequences. Our
proposed criterion managed in identifying the closest model order. In the
\cref{fig:aug-sys30}, apart from MDL, other polylines converged to the left-down
corner. Consequently, those model selection criteria were prone to choose a model
with much less complexity and failed for detecting the appropriate model order.

\section{Conclusion}
\label{sec:conclusion}

In this paper, we have proposed an approach for inferring time-invariant linear
dynamical system from data which is able to identify the latent
dimension in an automatic way. The proposed algorithm also avoids the drawback
of a standard likelihood-based EM algorithm: possible convergence to the
boundary of the parameter space. The criterion proposed in this paper could be
extended in many ways. In our algorithm, we have seen an example that it is
used on NARMA sequences. 

Instead of just using MDL as a model selection criterion to select one promising
model from a set of alternatives, we also integrate this criterion directly in
the EM algorithm. This allows practitioners to perform redundant state variable
annihilation, as we have done in our experiment. In this way, we seamlessly implement
model estimation and model selection in a single algorithm. Experimental results
demonstrated the decent performance of our proposed algorithm on model selection.

\section*{Appendix}

	We start from a Bayesian treatment of the code length. Within the Bayesian formalism, we have
	\begin{align}
	\length(\hat{\theta},Y)  = -\log p(\hat{\theta},Y)
	\end{align}
	where we have omitted the ceiling function $ \lceil \cdot \rceil $ for brevity.
	
	By expanding the right side according to Bayesian rules, we get the description length for both the
	model and the observations under this model. While it is impossible to evaluate
	its value with the true but unknown parameters, we may approximate this quantity
	by the truncated ones. The total code length is written
	as:
	\begin{align}
	\begin{split}
	\length(\hat{\theta},Y) & \approx -\log p(\theta) - \log p(Y|\theta) \\
	& \approx -\log(\delta p(\hat{\theta})) - \log p(Y|\hat{\theta})
	\end{split}
	\end{align}
	
	To evaluate its value, the second term is expanded to second order on the true
	value $ \theta $, which is written as:
	\begin{align}
	\begin{split}
	-\log p(Y|\hat{\theta}) \approx -\log p(Y|\theta) - (\theta - \hat{\theta})^T \frac{\partial \log p(Y|\theta)}{\partial \theta} \\ 
	- \frac{1}{2} (\theta - \hat{\theta})^T F(Y|\theta) (\theta - \hat{\theta})
	\end{split}
	\end{align}
	where $ F_\theta (Y|\theta)= \big [\frac{\partial^2 \log p(Y|\theta)}{\partial
		\theta_i \partial \theta_j} \big ] \big |_{\theta = \hat{\theta}}$. We also
	write it as $ F_\theta $ when the arguments are not important to our analysis or
	are clear from the context. The subscript indicates which variables the derivative
	is taken on.
	
	We are now in a position to obtain the expected code length. Take the
	expectation on both sides, we obtain:
	\begin{align}
	\label{eq:para:theta}
	\begin{split}
	\E \length(\hat{\theta},Y) =& -\log \delta - \log p(\theta) \\
	& - \frac{1}{2}\E (\theta -\hat{\theta})^T F_\theta(Y,\theta) (\theta - \hat{\theta})
	\end{split}
	\end{align}
	where the first-order term in Taylor series vanishes because $ \hat{\theta} $ is
	an ML estimate and the assumption that the quantization error is distributed
	uniformly, so we also have $ \E [\theta - \hat{\theta}] = \bm{0} $.
	
	Taking expectation of the second term may be difficult. For the moment, we
leave it as it is. Combine the above results,and by smoothness, we get:
	
	\begin{align}
	\begin{split}
		\length(\hat{\theta},Y) &= - \ln \delta - \ln( p(\hat{\theta})) -\ln(Y|\theta) \\ 
	&\qquad- \frac{1}{2} \E (\theta - \hat{\theta})^T F_\theta(Y|\theta)(\theta - \hat{\theta})
	\end{split}
	\end{align}
	where $ F_\theta(Y|\theta) $ is given by 
	\begin{align}
	F_\theta(Y|\theta) = \frac{\partial^2}{\partial \theta_i \partial \theta_j} \big [p(Y,\hat{X}|\theta)\big ] - \frac{1}{2} \ln{\det F_x(Y,\hat{X}|\theta)}
	\end{align}
	
	However, we know that apart from the quantization of parameters, another issue
that needs special attention is the latent variables. As opposed to common
static parametric models, the latent variables in LDS need careful processing.
In an LDS, each observation is linearly correlated with an underlying latent
state. LDS captures the dynamics via the evolving latent variables over the
time course. Identifying the evolving process of the latent state helps
practitioners discover the dynamical patterns behind observations.
	
From the standard viewpoint of MDL, the latent states are in fact
a part of model and can be recovered after we get the observations and model
parameters. In other words, the latent variables can be inferred from data and
are not totally independent. Conversely,
after recovering the hidden states under quantized parameters and observations,
they in turn can determinate the data in an implicitly way. To avoid getting
stuck in this dilemma, we attempt to isolate two different effects so as to 
investigate the latent variables under quantized parameters.

	We start from \cref{eq:para:theta}. Let $ \hat{X} $ be the collection of latent
	states that we gather after receiving the signals. As standard in the EM algorithm,
	the latent states $ \hat{X} $ are iteratively updated and the one maximizes $
	p(Y|\hat{\theta}) $ are valued high. In other words, $\hat{X}$ is the ML
	estimate for $ \ln p(Y|\hat{\theta}) $.
	
	The likelihood proceeds by integrating out the latent variables $ \hat{X} $ to
	find the probability of observing $ Y $ under the parameter $ \theta $.
	\begin{align}
	\label{eq:lik:x}
	\log p(Y|\theta) &= \log \int p(Y,X|\theta)dX 
	\end{align}
	
	The minimal code length under true parameters is given as:
	\begin{align}
	\label{eq:code:y}
	\length p(Y|\theta) &= -\log \int_X p(Y|X,\theta) p(X,\theta)dX
	\end{align}
	
	Suppose that the posterior is highly peaked, we approximate the integrand with
	Laplace's method, which takes the Taylor series around an ML estimate $ \hat{X}
	$ up to second order. Which is expressed as:
	\begin{align}
	p(Y,X |\theta) \approx p(Y,X|\theta) - \frac{1}{2} (X-\hat{X})^T F_X(Y,X|\theta) (X - \hat{X})
	\end{align}
	where the first-order term vanishes because in each step of the EM algorithm, $
	\hat{X} $ is updated by setting derivative of $ p(Y,X|\theta) $ to zero; the matrix $ F_X(Y|\theta) = \left[ - \dfrac{\partial^2 p(Y,X|\theta) }{\partial \bm{x}_i \partial \bm{x}_j} \right]\bigg |_{X=\hat{X}} $.
	
	By substituting this approximation for the integrand in \cref{eq:lik:x}, we obtain:
	\begin{align}
	\label{eq:lik:app}
	\begin{split}
	&p(Y|\theta)\\
	 = & \exp \big [p(Y,\hat{X}|\theta)\big ] \int_X \exp\big [-\frac{1}{2}(X-\hat{X})^T F_X (X-\hat{X})\big ]dX \\
	 = &\exp\big [p(Y,\hat{X}|\theta)\big ]\frac{(2\pi)^d}{\sqrt{\det F_x}}
	\end{split}
	\end{align}
	
	In terms of logarithms, the above result can be readily converted into:
	\begin{align}
	\label{eq:lik}
	\log p(Y|\theta) = \log p(Y|\hat{X},\theta) + \frac{d}{2}\log 2\pi -\frac{1}{2} \log \det F_{X}
	\end{align}
	
	
	Having obtained the approximation for likelihood function, we return to
	computing the expectation of total code length. For discrete parameters like
	model order, its code length can be obtained without considering quantization, just
	$ \lceil-\ln p(d) \rceil $. But for other parameters in $ \theta $ whose values are in a real
	domain, the truncation becomes necessary and inevitable for practical
	applications. Different from truncation real-valued observations, the
	truncation of parameters bring much difficulty on analysis.
	
	To compute the total code length, it is required to calculate the expectation of
	the following term:
	\begin{align}
	\label{eq:corr:lik}
	- \frac{1}{2} \E (\theta - \hat{\theta})^T F_\theta (\theta - \hat{\theta})
	\end{align}
	where $ F_\theta = \big [ \dfrac{\partial^2 \log p(Y|\theta)}{\partial \theta_i \partial \theta_j}\big ]\big|_{\theta = \hat{\theta}}$.
	
	Substitute the approximation \cref{eq:lik} into \cref{eq:corr:lik}, yielding:
	\begin{align}
	-\frac{1}{2} \E (\theta - \hat{\theta})^T F_{\theta,X} (\theta - \hat{\theta})
	\end{align}
	where $ F_{\theta,X} = \frac{\partial^2}{\partial \theta_i \partial \theta_j} \big\{ \log p(Y|\hat{X},\theta) - \frac{1}{2}\log \det F_X \big \} $

	
	To evaluate this formula, we follow \cite{Lanterman2001} for a closed-form result for
	multivariate parameter vector. According to \cite{Lanterman2001}, we make a change to a new
	coordinate system such that the inner product could be more conveniently
	processed. Suppose the variables corresponding to $ \theta_i $ and $ \theta_j $
	in a new coordinate could be respectively represented as $ z_i $ and $ z_j $. The
	conversion from $ \theta_i $ to $ z_i $ is chosen following our initiative, so
	that $ \E \{\theta^T F_{\theta,X} \theta\}   =  \E\{\bm{z}^T \bm{z} \}$. 
	
	To this end, we take matrix decomposition for $ F_{X,\theta} $, which can be
	easily shown as a symmetric and positive matrix.
	\begin{align}
	F_{X,\theta} = U \Lambda U
	\end{align}
	
	It is straightforward to show that the transform takes the form of $ \theta =
	U^{-1} \Lambda^{-\frac{1}{2}} \bm{z}$, where $ U^{-1} $ is an inverse to $ U
	$, $ UU^{-1} = I $; $ \Lambda^{-\frac{1}{2}} $ is a diagonal matrix
	which stacks reciprocals of the square roots of eigenvalues of $ \Lambda $ as
	the major diagonal. A simple calculation reveals that, this transformation is
	indeed what we intended for.
	\begin{align}
	\theta^T F_{X,\theta} \theta = \theta^T U \Lambda^{\frac{1}{2}} \Lambda^{\frac{1}{2}} U \theta = \bm{z}^T \bm{z}
	\end{align}
	where we have used the fact that $ \bm{z} = U\Lambda^{\frac{1}{2}} $.
	
	The density in new coordinate system, which we will denote with $ p_z(\cdot) $,
	correlates the original one through a Jacobian matrix.
	\begin{align}
	\label{eq:jacob}
	Jacob = \det \big (\frac{\partial \bm{z}}{\partial \theta} \big ) = \det(U^{-1}\Lambda^{-\frac{1}{2}}) \overset{(a)}{=} Jacob(\Lambda^{-\frac{1}{2}}) \overset{(b)}{=} \sqrt{\det F_{X,\theta}}
	\end{align}
	where $ (a) $ is by the definition of unitary matrix; $ (b) $ arises from
	operating the eigenvalues on the diagonal, $ \det(\Lambda) = \prod_i \sqrt{\lambda_i } = \sqrt{\prod \lambda_i} = \sqrt{\det F_{X,\theta}}$.
	
	So we have the relation on variables between the two coordinates:
	\begin{align}
	\label{eq:transm:jacob}
	p_z(\bm{z}) = \frac{p(\theta)}{Jacob}= \frac{p(\theta)}{\sqrt{F_{X,\theta}}}
	\end{align} 
	
	In our application, we seek to minimize the maximum possible error. Following
	\cite{Lanterman2001,Wallace1987,Wallace1992}, we consider quantizing this multivariate parameter vector using
	\emph{optimum quantizing lattices}, which, in two dimensional space, appears as a
	hexagonal grid. In three dimensional space, forms a cubic lattice. We omit the
	proof, and directly present the results here. If the quantization is performed in
	the way of \emph{optimum quantizing lattice}, we can get a closed-form expression for the
	expectation of \cref{eq:corr:lik} as follows:
	\begin{align}
	\label{eq:transform:prob}
	\E [(\bm{z}-\hat{\bm{z}})^T (\bm{z}-\hat{\bm{z}})] = d \kappa_d \delta^{\frac{2}{d}}
	\end{align}
	where $ \kappa_d $ is a constant relating to the geometric property of lattice.
Although an exact value for $ \kappa_d $ is not readily known, we can get an
approximation by using its upper and lower bound $
\frac{\Gamma(\frac{2}{d})\Gamma(\frac{2}{d}+1) }{d\pi} > \kappa_d
>\Gamma(\frac{d}{2} +1)^\frac{2}{d}(d+2)\pi$ \cite{Zador1982,Wallace1987}. A
good property is that when $ d $ grows, the gap between the upper and lower bound
becomes tight. $ \kappa_d $ approaches its asymptotic value as $ d $ increases,
$ \kappa_d \rightarrow \frac{1}{2\pi e}  $ \cite{Conway2013}. As $ \kappa_d $
changes slowly, \cite{Figueiredo2002} proposes to approximate it with $
\frac{1}{12} $. In our study, we use its asymptotic value instead. This choice
has no effects on the minimization. 
	
	Carrying out quantization in new coordinate is also necessary as we are seeking
	to minimize the maximum possible error that is induced from the truncation. To avoid
	confusion, we will add subscript $ z $ to the density function $ p $ in this new
	coordinate.
	\begin{align}
	\begin{split}
	&\E [- \log \delta_z p_z( \bm{z}) - \log p_z(Y|\hat{\bm{z}}) ] \\
	&\approx -\log \delta_z - \log p_z(\bm{z}) - \log p_z(Y|\bm{z}) + \frac{d}{2} \kappa_d \delta_z^{\frac{2}{d}}
	\end{split}
	\end{align}
	
	Straightforward calculus reveals that by setting $ \delta_z =
	\kappa^{-\frac{d}{2}} $, we can get the maximum possible error with regards to
	the quantization. 
	\begin{align}
	\frac{d}{2} \log \kappa_d  - \log p_z(\bm{z}) - \log p_z(Y|\bm{z}) + \frac{d}{2}
	\end{align}
	
	Transforming back to original coordinate using \cref{eq:transm:jacob}, we know
	immediate that the total message length can be expressed as:
	\begin{align}
	\label{eq:min:obj}
	\begin{split}
	&-p(\theta) + \frac{1}{2} \log \det F_{X,\theta} - \log p(Y|\theta) + \frac{d}{2}(1+\log \kappa_d) \\
	\overset{(a)}{\approx}& -p(\theta) + \frac{1}{2} \log \det F_{X,\theta} + \frac{d}{2}(1+\log \kappa_d) \\
	& - \log p(Y|\hat{X},\theta) - \frac{d}{2}\log 2\pi +\frac{1}{2} \log \det F_{X} \\
	= & -p(\theta) - \log p(Y|\hat{X},\theta)  + \frac{1}{2} \log \det F_{X,\theta} F_X \\
	&+ \frac{d}{2}(1 + \log \kappa_d - \log 2 \pi) \\
	& \rightarrow  -p(\theta) - \log p(Y|\hat{X},\theta) + \frac{1}{2} \log \det{F_{X,\theta} F_{X}} - d \log{2 \pi}
	\end{split}
	\end{align}
	where $ (a) $ is established by substituting the approximation \cref{eq:lik} into $ \log p(Y|\theta) $. 
	
	Therefore, the particular form of the MDL leads to the following estimation criterion.
	\begin{align}
	\begin{split}
	\{\hat{\theta}, d\} & = \argmin_{\theta,d} \big \{ -p(\theta) - \log p(Y|\hat{X},\theta) \\
	& + \frac{1}{2} \log \det F_{X,\theta} F_X + \frac{d}{2}(1 + \log \kappa_d - \log 2 \pi) \big \}
	\end{split}
	\end{align}
	
	The remaining work left to us is to estimate the $ F_X $ and $ F_{X,\theta} $.
	For general settings, the two matrices cannot be obtained analytically. To deal
	with this difficulty, we propose to estimate these two key components from
	observations. We replace $ F_{X, \theta} $ and $ F_{X} $ with approximation from
	complete data, which help us reduce the bias that may be inevitable for single
	sequence and the possible variance for small data.
	
	Suppose that the sequences $ \{\bm{y}_1,\cdots,\bm{y}_n \} $ are generated
	\emph{i.i.d.} from unknown density $ p(\bm{y}) $, so we have the likelihood:
	\begin{align*}
	\log p(Y|\theta) &= \sum_i \log p(\bm{y}_i|\theta)\\
	\log p(Y,\hat{X}|\theta) &= \sum_i \log p(\bm{y}_i,\hat{\bm{x}}_i|\theta)
	\end{align*}
	The empirical Fisher information is estimated to be:
	\begin{align*}
	F_\theta(Y,\hat{X}|\theta) = \sum_i F_\theta(\bm{y}_i,\hat{\bm{x}}_i|\theta)
	\end{align*}
	
	Its asymptotic behavior can be established through taking expectation of $
	F_\theta $. From the \emph{i.i.d.} assumption, as $ N\rightarrow \infty $, the
	summand asymptotically converges to the value of $ N $ times expected value $
	F_\theta $ on a particular sample in the sense of expected least square.
	\begin{align}
	\label{eq:large:approx}
	F_\theta(Y,X|\theta) \approx N \E_{p(\bm{y}|\hat{\bm{x}},\theta)} F_\theta(\bm{y},\hat{\bm{x}}|\theta)
	\end{align}
	Under a large sample size, $ F_\theta(Y,\hat{X}) $ can thus be approximated through:
	\begin{align}
	\label{eq:ftheta}
	\begin{split}
	&\frac{1}{2} \log\det F_\theta(Y,X|\theta) \\
	&\overset{(a)}{\approx }\frac{1}{2}\log \det \{ N \E_{p(\bm{y},\bm{x}|\theta)} F_\theta(Y,X|\theta) \} \\
	&= \frac{1}{2} \log \det N I_d + \frac{1}{2}\log \det \E_{p(\bm{y},\bm{x}|\theta)} F_\theta(\bm{y}_1,\bm{x}_1|\theta) \} \\
	&\overset{(b)}{=} \frac{d}{2}\log N + \E_{p(\bm{y},\bm{x}|\theta)} F_\theta(\bm{y}_1,\bm{x}_1|\theta)\\
	&\rightarrow \frac{d}{2}\log N
	\end{split}
	\end{align}
	where $ I_d $ is an identity matrix of size $ d\times d$, and the $ (a) $ holds
	by putting \cref{eq:large:approx} into the formula. The $ (b) $ is due to the
	\emph{i.i.d.} assumption on the data and the basic properties of the determinant.  A
	similar result can be established for $ F_{X,\theta} $. Therefore, the
	logarithmic form of $ p(Y) $ can be asymptotically approximated by:
	\begin{align*}
	&-p(\theta) - \log p(Y|\hat{X},\theta)  + \frac{d}{2}(1 + \log \kappa_d - \log 2 \pi + 2 \log N) \\
	&\quad +\frac{1}{2} \log \det \E_{p(\bm{y}|\bm{x},\theta)} F_{X,\theta} + \frac{1}{2} \log \det \E_{p(\bm{y}|\bm{x},\theta)} F_X \\
	&\overset{(a)}{\longrightarrow}-\log p(Y|\hat{X},\theta) + d\log N
	\end{align*}
	where $ (a) $ holds only when $ N\rightarrow \infty $. Hence, when the sample
	size approaches infinity, we recover the BIC criterion as a special case
	except for a constant coefficient. 
	
	Remarkably, for HMM whose latent variables are in the discrete domain, there is no
	loss in quantization on latent variables. We simply drop the term associated
	with the derivatives in terms of $ X $. 
	
	In the remaining study, we demonstrate how $ F_{\theta} $, $ F_{X} $, and $
	F_{\theta,X} $ can be obtained or approximated. To do so, we first present some results on the posterior probabilities
	of observations at time $ t $.
	\begin{align}
	\E_{p(\bm{y}_t|\bm{x}_t)}[\bm{y}_t|\bm{x}_t] &= C \bm{x}_t \\
	\E_{p(\bm{y}_t|\bm{x}_{t-1})}[\bm{x}_t|\bm{x}_{t-1}] &= A \bm{x}_{t-1}
	\end{align}
	
	We continue in this way and obtain:
	\begin{align}
	\E [\bm{y}_t|\bm{x}_{t-m}] = C A^{m-1} \bm{x}_{t-m}
	\end{align}
	
	While parameters are time-invariant, the derivatives with respect to them do
not involve special techniques. In contrast, the derivative in terms of latent
variables needs careful treatment. A critical distinction between latent
variables and parameters is that, the former is often attached with a time
stamp. Suppose we are dealing with a stable dynamical system, the inference of
a latent variable in fact degenerates over time. Thus, the derivative with
respect to latent variables should take this point into account.
	
	We unfold the LDS and see how the latent variables relate to the observations.
	The original form can be equivalently reformed as:
	\begin{align}
	\bm{y}_{t+1} = CA\bm{x}_t + C \bm{w}_t + \bm{v}_{t+1}
	\end{align}
	where $ \bm{w}_{t} $ and $ \bm{v}_{t+1} $ absorb the disturbances or errors
	generated during model fitting. 
	
	We continue in this way and obtain:
	\begin{align*}
	\bm{y}_{t} &= CA\bm{x}_{t-1} + C \bm{w}_{t-1} + \bm{v}_{t} \\
	&= \cdots = CA^m\bm{x}_{t-m} + CA^{m-1}\bm{w}_{t-m} +\cdots+ C \bm{w}_t + \bm{v}_{t+1}
	\end{align*}
	where $ \bm{w} $ and $ \bm{v} $ capture the disturbances
	existing along the process, and are assumed to be independent of both
	observations and states. We can also summarize the above continued equation concisely in a distribution:
	\begin{align}
	\bm{y}_t \sim \mathcal{N}(CA^m\bm{x}_{t-m},R_2) 
	\end{align}
	where $ R_2 $ is the covariance matrix for the observations.
	
	However, as we are dealing with the stable system, the accumulation of disturbances
	is limited in propagating along the model process. The statistical properties state that, for
	a stable LDS, the latent states are assumed to have a stable and time-invariant covariance matrix, which
	is denoted as $ R_1 $. This is a quite mild assumption yet can lead to a
	mathematically simplified analysis.
	
	Based on the assumptions, we have 
	\begin{align}
	\label{eq:fx}
	\begin{split}
	[F_{X}]_{k,l} &=  \sum_{i}\E \frac{\partial^2}{\partial x_i^l \partial x_i^k}
	p(\bm{y}_i,\bm{x}_i|\theta) \\
	&= N \E \frac{\partial^2}{\partial x_i^k \partial x_i^l}
	p(\bm{y}_i|\bm{x}_i) p(\bm{x}_i) \big |_{\bm{x} = \hat{\bm{x}}_i}\\ 
	& = N \E \sum_m \frac{\partial^2}{\partial x_i^k \partial x_i^l} p(\bm{y}_i, \bm{x}_{i-m}) \\
	& \overset{(a)}{\leq} N \E \max_m \big \{ \frac{\partial^2}{\partial x_i^k \partial x_i^l} p(\bm{y}_i | \bm{x}_{i-m} )  + \frac{\partial^2}{\partial x_i^k \partial  x_i^l} p(\bm{x}_{t-m})  \big \} \\
	&\overset{(b)}{=} N \sum_m 2 [C A^m R_2 (CA^m)^T +   R_1]_{k,l}\\
	&\overset{(c)}{\leq} 2 N [(CQC^T) + R_1]_{k,l}  
	\end{split}
	\end{align}
	where $ (a) $ holds because of the chain rule of derivative and the inequality
	holds due to the fact that probabilities are less than one on magnitude. $
	\bm{x}_{t-m} $ is the $ m $-lagged latent variables. As we are minimizing the
	maximum possible description length, we take maximum on the right side. Equality
	$ (b) $ is established by taking expectations with respect to $ \bm{y}_t $ and $
	\bm{x}_t $. In inequality $ (c) $, we notice that the first term in summand is
	positive semidefinite (PSD) matrix. It follows that $ \sum_m A^m \Sigma (A^m)^T <
	\sum_{m = 1}^\infty A^m \Sigma (A^m)^T  $. This infinite summation can be shown
	as the solution to an equation $ A X A^T - X = R_2 $, whose solution is
	denoted as $ Q $.

	The equation is more referred as discrete Lyapunov equation in the subject of
	automation, which is the main tool in studying the stability and its asymptotic
	behavior. The $ Q $ reflects how the model trace reacts to a minor disturbance
	on observations and whether a system could be recovered from a systematic
	identification error. Hence, its appearance here is not supervising. Since by
	assumption, we are dealing with a stable LDS, this equation is ensured to have
	PSD matrix as a solution. Otherwise, when the system is unstable, there are
	cases where any small errors on quantization of latent variables will result in
	significant deviation from the true model trace.

	The PSD matrix $ Q $, together with $ R_1 $ constitute a meaningful
	quantity reflecting how much degree the errors on quantization could affect
	system outputs, in particular, errors and disturbances caused by
	misspecification of model order and the quantization. If any term in the summand
	is large, it means that an LDS is either too sensitive to quantification, or the
	latent variables are unstable enough and have a great chance to be far from the
	true ones. In either case, we need a large description length for this model.
	
	Substitute \cref{eq:ftheta} and \cref{eq:fx} into last equation of \cref{eq:min:obj}, yielding
	\begin{align}
	\begin{split}
	\{\hat{\theta}, d\}  &= \argmin_{\theta,d} \big \{ -p(\theta) - \log p(Y|\hat{X},\theta)  \\
	&+\frac{1}{2} \log \det [C Q C^T +R_1]  + \frac{d}{2} \log \frac{2 N^2}{(2 \pi)^2}  \big \}
	\end{split}
	\end{align}
	where we approximate the $ \kappa_d $ using its asymptotic value $ \frac{1}{2 \pi e} $.
	
	For ML estimates of parameters, we drop the first term in the summand and leave
	others unchanged. The final model order is selected to be the one which
	minimizes the above objective function on the ML estimates of parameters, $
	\theta_{ML} $. Apart from description length expressed with $ p(Y|\theta) $,
	every increase on the model order will incur a cost that at a logarithmic scale
	with $ N $. This increase is worthwhile only if it may bring up the robustness
	of LDS to latent variables, which is reflected in the log-determinant term.

\bibliographystyle{IEEEtran}
\bibliography{bibtex}

\end{document}